\documentclass{article} % For LaTeX2e
\usepackage[table,xcdraw]{xcolor}
\usepackage[preprint]{neurips_for_arxiv}

\usepackage{amssymb,amsfonts,bm,amsthm,stackengine}
\setstackEOL{\\}
\usepackage{mathtools}
\newcommand{\smallmodel}{\small SMALL-MODEL\normalsize}
\newcommand{\mg}{\small MAG\normalsize}
\newcommand{\snip}{\small SNIP\normalsize}
\newcommand{\synflow}{\small SYNFLOW\normalsize}
\newcommand{\rand}{\small RAND\normalsize}
\newcommand{\grasp}{\small GRASP\normalsize}
\newcommand{\ltr}{\small LTR\normalsize}
\newcommand{\rps}{\small RPS\normalsize}
\newcommand{\roast}{\small ROAST\normalsize}
\newcommand{\roastpp}{\small STABLE{-}RPS\normalsize}
\newcommand{\roastppsmall}{\tiny STABLE{-}RPS\normalsize}
\newcommand{\robe}{\small ROBE\normalsize}
\newcommand{\randomfold}{\small RANDOMFOLD\normalsize}
\newtheorem{theorem}{Theorem}

\def\eqref#1{equation~\ref{#1}}
\def\1{\bm{1}}

\DeclareMathAlphabet{\mathsfit}{\encodingdefault}{\sfdefault}{m}{sl}
\SetMathAlphabet{\mathsfit}{bold}{\encodingdefault}{\sfdefault}{bx}{n}

\usepackage{hyperref}
\usepackage{url}
\usepackage{subcaption}
\usepackage{graphicx}
\usepackage{multirow}
\usepackage{enumitem}

\title{In defense of parameter sharing for model-compression}

\author{Aditya Desai \\
Department of Computer Science\\
Rice University\\
Houston, TX, USA \\
\texttt{apd10@rice.edu} \\
\And
Anshumali Shrivastava \\
Department of Computer Science \\
Rice University \& ThirdAI Corp\\
Houston, TX, USA \\
\texttt{as143@rice.edu} \\
}

\newcommand{\ssection}[1]{ \section{#1}}
\newcommand{\ssubsection}[1]{ \subsection{#1}}

\begin{document}

\maketitle
\begin{abstract}
When considering a model architecture, there are several ways to reduce its memory footprint. Historically, popular approaches included selecting smaller architectures and creating sparse networks through pruning. More recently, randomized parameter-sharing (RPS) methods have gained traction for model compression at start of training. In this paper, we comprehensively assess the trade-off between memory and accuracy across RPS, pruning techniques, and building smaller models. Our findings demonstrate that RPS, which is both data and model-agnostic, consistently outperforms/matches smaller models and all moderately informed pruning strategies, such as  \mg, \snip, \synflow, and \grasp, across the entire compression range. This advantage becomes particularly pronounced in higher compression scenarios. Notably, even when compared to highly informed pruning techniques like Lottery Ticket Rewinding (\ltr), RPS exhibits superior performance in high compression settings. This points out inherent capacity advantage that RPS enjoys over sparse models. Theoretically, we establish RPS as a superior technique in terms of  memory-efficient representation when compared to pruning for linear models. This paper argues in favor of paradigm shift towards RPS based models. During our rigorous evaluation of RPS, we identified issues in the state-of-the-art RPS technique \roast, specifically regarding stability (\roast's sensitivity to initialization hyperparameters, often leading to divergence) and Pareto-continuity (\roast's inability to recover the accuracy of the original model at zero compression). We provably address both of these issues. We refer to the modified RPS, which incorporates our improvements, as \roastpp.

\end{abstract}
\vspace{-0.2cm}
\ssection{Introduction}
An essential question in model design is that \textit{ given a memory budget for parameters, what is the best architecture one can create for a given task}. The solution would require us to consider the model's capacity, learning-friendliness, and generalization capability. Limited understanding of these aspects makes the problem hard. Traditionally, model architectures are designed with task-domain expertise, often augmented with neural architectural search \citep{Pham2018_ilh}. However, recently, it was shown by Lottery ticket hypothesis \citep{frankle2018lottery}(LTH) and follow-up works \citep{frankle2020pruning, pmlr-v119-frankle20a} that given a model architecture, the model architecture itself can be sparsified to obtain lighter models. If done intelligently, this sparsification can often maintain the accuracy to certain levels of sparsity. This has drawn interest to another exciting question \textit{given a memory budget and an architecture, what is the best model we can derive from the architecture.}. While pruning was traditionally applied as a post-processing operation \citep{mozer1988skeletonization, lecun1990advances,hassibi1992second, han2015learning}, LTH and follow-up works show that global pruning is an excellent technique to reduce the memory requirement of the model at the start of training. Since then, many works to reduce the memory footprint of the model at the beginning of training have focused on pruning as a critical technique of the algorithm \citep{lee2018snip,tanaka2020pruning,zhang2022grasp,evci2020rigging,prospr}. In this paper, we raise the question \textit{``Is pruning the correct method to reduce model parameters?".}

We suspect that pruning parameters to reduce its memory footprint is a harsh operation affecting the model capacity adversely. A simple illustration highlights this issue. Consider pruning a $n \times d$ embedding table. If we prune beyond $d\times$, we start getting degenerate zero embeddings. The same problem is present in pruning models with other components such as linear or convolutions. As we prune more and more parameters, we start dropping nodes and filters, which would affect model capacity. In the case of embedding tables, recent results from parameter sharing based compression\citep{hashtrick, desai2023hardware,MLSYS2022_1eb34d66} seems promising where 100GB sized embeddings were reduced to 10MB without loss of quality \citep{NEURIPS2022_dbae9151}. In this paper, we evaluate parameter sharing as an alternative to reduce the memory footprint of the model. In our embedding table example, randomized parameter sharing (\rps) is guaranteed to give non-degenerate embeddings no matter how small the memory is and unique embeddings with high probability. In the paper, we theoretically show that parameter sharing has a better memory capacity (thus accuracy) tradeoff in arguably a simple setting of linear models. We also rigorously evaluate memory-accuracy tradeoffs of RPS, which is model, initialization, and data agnostic, against various uninformed, moderately informed, and highly informed pruning strategies.

We find that {\rps} not only outperforms {\rand} but also consistently outperforms/matches moderately informed pruning methods  such as {\mg}, {\snip}\citep{lee2018snip}, {\grasp}\citep{zhang2022grasp} and {\synflow}\citep{tanaka2020pruning}, especially in high compression regimes (it is competitive in low-compression region to best out of all pruning methods). While {\rps} is inferior to the highly informed and computationally expensive pruning technique of Lottery Ticket Rewinding (\ltr) \citep{pmlr-v119-frankle20a} in low compression region, it surprisingly is significantly better than {\ltr } in high compression region. This highlights {\rps}'s inherent capacity advantage compared to pruning. Due to this capacity disadvantage, even a highly informed pruning (like {\ltr}) is worse than a vanilla uninformed parameter-sharing method. While we restrict our attention to {\rps}, which is the model, initialization, and data-agnostic method, our results encourage research in informed parameter-sharing-based models to achieve even better memory-accuracy tradeoff.

In our endeavor to rigorously evaluate {\rps }, we use the SOTA {\rps} scheme of {\roast } \citep{desai2023hardware} and find two significant issues in {\roast } (1) stability: The {\roast } with its global-memory-sharing introduces a new hyper-parameter of initialization standard-deviation of the {\roast } array. This standard deviation determines the different multiplicative factors used for each component. We find it essential to choose a standard deviation that keeps the multiplicative factors in check. Any extreme choice, such as a low standard deviation, will blow up multiplicative factors, causing divergence, and a high standard deviation will suppress multiplicative factors, causing slow learning. We provide a gradient scaling scheme that provably improves the stability of the training and removes its sensitivity to initialization standard deviation. (2) Pareto-continuity: {\roast } cannot recover the accuracy of the entire model when compression is set to $1\times$ (no compression). This is a desirable property for any compression method, and pruning achieves this naturally. We propose a different mapping function that maintains the cache-efficient memory coalescing, obtains the optimal number of collisions, and also has the Pareto-continuity property - when compression is set to $1\times$, there is no collision; thus, the model with {\rps } is equivalent to the entire model. {\roast }, along with improvements in gradient updates and mapping function, is collectively referred to as {\roastpp} in the paper. We want to stress that without the mapping scheme used in {\roastpp}, the theoretical results of the superiority of parameter sharing over pruning would not have been possible.

We make the following contributions,
\begin{itemize}[nosep, leftmargin=*]
\item Identify and elucidate the shortcomings of the existing state-of-the-art parameter-sharing method {\roast } about issues related to stability and Pareto-continuity.

\item Introduce a set of enhancements that provably and demonstrably rectify the stability and Pareto-continuity problems within {\roast}. These enhancements are collectively referred to as {\roastpp}.

\item We conduct a comprehensive evaluation of {\roastpp } in comparison to building small models and pruning at initialization methods such as (1) uninformed pruning, (2) Moderately informed pruning, and (3) highly informed pruning. Our findings establish that uninformed parameter sharing is already competitive/superior to moderately informed pruning strategies at all compression factors and highly informed pruning at higher compression.

\item We provide theoretical insights to explain why compressed models based on parameter sharing exhibit superior capacity compared to pruned models.

\end{itemize}

\ssection{Background}
There are various methods for model memory footprint reduction, such as designing smaller models \citep{iandola2016squeezenet, howard2017mobilenets, prabhu2018deep}, low-rank component models \citep{jaderberg2014speeding, novikov2015tensorizing}, parameter sharing based components \citep{hashtrick, MLSYS2022_1eb34d66,desai2023hardware,desai2021semantically, NEURIPS2022_dbae9151} and sparse network \citep{spring2017scalable,chen2020slide,mocanu2018scalable,bellec2017deep}. In this work, we focus on the problem of deriving a model architecture from a given full architecture, explicitly focusing on pruning and parameter sharing. 
\begin{figure}
    \centering
    \includegraphics[scale=0.25]{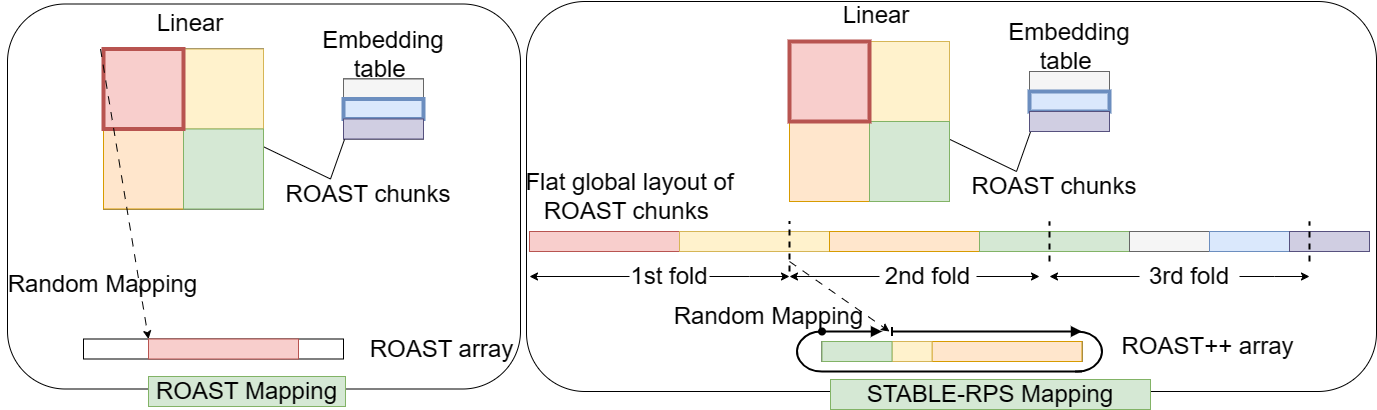}
    \caption{{\roast} and {\roastpp} mapping functions. While {\roast} maps chunks independently, {\roastpp} essentially flattens the blocks into a single parameter array, divides it into partitions of size $m$, and independently hashes each partition. }
    \label{fig:expmapping}
    %\vspace{-0.5cm}
\end{figure}
\ssubsection{Parameter sharing based methods and {\roast}}
The parameter-sharing-based methods maintain a small repository of parameters, and all the weights are drawn from this repository of weights using two hash functions. Let the total size of the model $\theta$ be $|\theta| = n$ and size of repository $\psi$ be $|\psi| = m < n$, the two hash functions generally used are $h: [n] \rightarrow [m]$ and $g: [n] \rightarrow \{ \pm 1\}$ where [n] = $\{0, ..., n-1\}$. The exact hash function inputs might vary depending on the exact algorithm. The recovered weight with global index $i$ under element-wise mapping used in HashedNet \citep{hashtrick} is

\begin{equation}
    g:[n] \rightarrow \pm 1,\; h:\mathbf{N} \rightarrow [m] \quad \theta[i] = g(i) \psi[h(i)] \tag{element-wise}
\end{equation}
The recovered weight is then used in the computation defined by the full model. Due to the independent mapping of adjacent elements, this mapping suffers from high cache misses. This issue was first fixed for embedding tables in \citet{MLSYS2022_1eb34d66} using {\robe}-Z hashing.
\begin{equation}
    g:[n] \rightarrow \pm 1,\; h:\mathbf{N} \rightarrow [m] \quad \theta[i] = g(i) \psi[(h(i//Z) + i \% Z ) \% m] \tag{\robe-Z}
\end{equation}
{\roast}\citep{desai2023hardware} extends this idea of {\robe}-Z to other modules, such as matrix multiplication, and proposes to use different shapes and sizes of chunks depending on the layer. {\roast} also introduced global memory sharing, which requires one to appropriately scale the recovered parameter to match the module's initialization scale. For instance, a parameter shared between two weights from two different matrices should be scaled differently for two matrices. Thus, {\roast} has scaling factors for each module. For ease of expression, we represent them as $\{\lambda_i\}_{i=1}^n$ for each weight. The recovered weight in {\roast} is,
\begin{equation}
    g:[n] \rightarrow \pm 1,\; h:\mathbf{N} \rightarrow [m] \quad \theta[i] = g(i) \lambda_i \psi[(h(\textrm{chunk-id}) + \textrm{chunk-offset}) \% m] \tag{\roast}
\end{equation}
chunk-id is the unique integer that identifies the chunk, and chunk-offset is the offset of the parameter inside the chunk. The mapping in {\roast} is illustrated in figure \ref{fig:expmapping}.
\vspace{-0.25cm}
\ssubsection{Pruning methods}
While there are various kinds of pruning techniques - structured and unstructured, in terms of accuracy-memory tradeoff, global pruning has seen the best results \citep{frankle2020pruning}. The general recipe for (global) pruning is assigning a score to each parameter and then pruning the parameters with the lowest scores.  Pruning has traditionally been applied as a post-training model compression technique. However, the Lottery ticket hypothesis (LTH) used iterative magnitude pruning (IMP) to find sparse sub-networks of randomly initialized models that can achieve high accuracy. Since then, pruning for training-cost reduction has been an active area of research. We have seen progress in LTH-based methods such as Lottery Ticket Rewinding (\ltr), which, instead of re-initializing the weights to the initial weights, rewinds the weights to a later point in training \citep{pmlr-v119-frankle20a}. Also, a variety of single-shot pruning strategies have evolved. Most pruning strategies differ in the scoring mechanism. Some standard baselines are  {\rand}:  randomly assigns the score and {\mg}: assigns a score based on the absolute value of the parameter. Single-shot Network Pruning based on Connection Sensitivity (\snip) \citep{lee2018snip} uses $|\frac{\partial \mathcal{L}}{\partial \theta} \circ \theta|$ for scoring the parameters whereas Gradient
Signal Preservation(\grasp)\citep{zhang2022grasp} uses $ H \frac{\partial \mathcal{L}}{\partial \theta} \circ \theta$. A random sample of training data is used for this scoring. Synaptic flow pruning (\synflow)\citep{tanaka2020pruning} uses a scoring mechanism purely based on the weights and thus does not use any data. There have been other methods such as \citet{prospr, evci2020rigging} and more. In this paper, we use {\mg}, {\snip}, {\synflow} and {\grasp} as a representative of moderately informed pruning and {\ltr} as representative of highly informed pruning.

\ssection{Stability of {\roastpp}} \label{sec:stability}
% highlight the issue in data.
This section discusses the stability issues with {\roast} and how to fix them. We start with empirical evidence of the sensitivity of {\roast} with the hyperparameter of the standard deviation of initialization used for {\roast} array ( henceforth called init-stdev). The Accuracy columns in Table~\ref{fig:stability} show that when we change the init-stdev from 1e-3 to 10, the convergence of {\roast} suffers at the two extremes. When the init-stdev is very low, the optimization diverges; when the init-stdev is very high, the optimization slows down. This is consistently observed across different sparsity levels and datasets. When we look at the norms of the solutions obtained during optimization, we observe that with lower init-stdev, the norms explode for most cases, indicating divergence, and with higher init-stdev, the norms are lower, meaning slow learning.
\begin{table}[t]
\caption{Sensitivity of {\roast} to init-stdev. (compression = $\frac{\mathrm{original}}{\mathrm{compressed}}$). {\roastpp} eliminates the sensitivity to init-stdev }\label{fig:stability}
%\vspace{-0.25cm}
\centering
\resizebox{0.9\textwidth}{!}{
\begin{tabular}{|cc|ccccccccccc|}
\hline
\multicolumn{1}{|c|}{}                               &                                   & \multicolumn{11}{c|}{\textbf{CIFAR-10(RESNET-20)}}                                                                                                                                                                                                                                                                                                                                                                                                                                                                                   \\ \hline
\multicolumn{2}{|c|}{\textbf{Compression}}                                               & \multicolumn{5}{c|}{\textbf{Accuracy}}                                                                                                                                                                                                                            & \multicolumn{1}{c|}{} & \multicolumn{5}{c|}{\textbf{Norm of Solution}}                                                                                                                                                                                           \\ \hline
\multicolumn{1}{|c|}{\textbf{}}                      & \textbf{init-stdev $\rightarrow$} & \multicolumn{1}{c|}{0.001}                        & \multicolumn{1}{c|}{0.01}                         & \multicolumn{1}{c|}{0.1}                          & \multicolumn{1}{c|}{1}                            & \multicolumn{1}{c|}{10}                           & \multicolumn{1}{c|}{} & \multicolumn{1}{c|}{0.001}                       & \multicolumn{1}{c|}{0.01}                         & \multicolumn{1}{c|}{0.1}                         & \multicolumn{1}{c|}{1}                           & 10                          \\ \hline
\multicolumn{1}{|c|}{}                               & \roast            & \multicolumn{1}{c|}{\cellcolor[HTML]{FF0000}10.0} & \multicolumn{1}{c|}{\cellcolor[HTML]{FFE800}86.9} & \multicolumn{1}{c|}{\cellcolor[HTML]{00FF00}91.5} & \multicolumn{1}{c|}{\cellcolor[HTML]{FFFE00}90.4} & \multicolumn{1}{c|}{\cellcolor[HTML]{FFB900}79.4} & \multicolumn{1}{c|}{} & \multicolumn{1}{c|}{\cellcolor[HTML]{FF0000}nan} & \multicolumn{1}{c|}{\cellcolor[HTML]{FF0000}639}  & \multicolumn{1}{c|}{\cellcolor[HTML]{05FA00}64}  & \multicolumn{1}{c|}{\cellcolor[HTML]{FFFF00}27}  & \cellcolor[HTML]{7CFF00}40  \\ \cline{2-13} 
\multicolumn{1}{|c|}{\multirow{-2}{*}{1.33$\times$}} & \roastpp         & \multicolumn{1}{c|}{\cellcolor[HTML]{00FF00}91.4} & \multicolumn{1}{c|}{\cellcolor[HTML]{00FF00}91.2} & \multicolumn{1}{c|}{\cellcolor[HTML]{00FF00}91.3} & \multicolumn{1}{c|}{\cellcolor[HTML]{00FF00}91.2} & \multicolumn{1}{c|}{\cellcolor[HTML]{00FF00}91.4} & \multicolumn{1}{c|}{} & \multicolumn{1}{c|}{\cellcolor[HTML]{01FE00}56}  & \multicolumn{1}{c|}{\cellcolor[HTML]{01FE00}56}   & \multicolumn{1}{c|}{\cellcolor[HTML]{01FE00}56}  & \multicolumn{1}{c|}{\cellcolor[HTML]{01FE00}56}  & \cellcolor[HTML]{01FE00}56  \\ \hline
\multicolumn{1}{|c|}{}                               & \roast            & \multicolumn{1}{c|}{\cellcolor[HTML]{FF0000}10.0} & \multicolumn{1}{c|}{\cellcolor[HTML]{FFE400}81.9} & \multicolumn{1}{c|}{\cellcolor[HTML]{00FF00}87.2} & \multicolumn{1}{c|}{\cellcolor[HTML]{08FF00}86.4} & \multicolumn{1}{c|}{\cellcolor[HTML]{FFC300}77.3} & \multicolumn{1}{c|}{} & \multicolumn{1}{c|}{\cellcolor[HTML]{FF0000}nan} & \multicolumn{1}{c|}{\cellcolor[HTML]{FF0000}1.1k} & \multicolumn{1}{c|}{\cellcolor[HTML]{1CE300}110} & \multicolumn{1}{c|}{\cellcolor[HTML]{ECFF00}29}  & \cellcolor[HTML]{7FFF00}40  \\ \cline{2-13} 
\multicolumn{1}{|c|}{\multirow{-2}{*}{10$\times$}}   & \roastpp         & \multicolumn{1}{c|}{\cellcolor[HTML]{00FF00}87.3} & \multicolumn{1}{c|}{\cellcolor[HTML]{00FF00}87.3} & \multicolumn{1}{c|}{\cellcolor[HTML]{00FF00}87.4} & \multicolumn{1}{c|}{\cellcolor[HTML]{00FF00}87.0} & \multicolumn{1}{c|}{\cellcolor[HTML]{00FF00}87.4} & \multicolumn{1}{c|}{} & \multicolumn{1}{c|}{\cellcolor[HTML]{01FF00}53}  & \multicolumn{1}{c|}{\cellcolor[HTML]{00FF00}53}   & \multicolumn{1}{c|}{\cellcolor[HTML]{00FF00}53}  & \multicolumn{1}{c|}{\cellcolor[HTML]{01FF00}53}  & \cellcolor[HTML]{01FF00}53  \\ \hline
\multicolumn{1}{|c|}{}                               & \roast            & \multicolumn{1}{c|}{\cellcolor[HTML]{FF0000}10.0} & \multicolumn{1}{c|}{\cellcolor[HTML]{FFF800}69.6} & \multicolumn{1}{c|}{\cellcolor[HTML]{00FF00}75.5} & \multicolumn{1}{c|}{\cellcolor[HTML]{25FF00}74.2} & \multicolumn{1}{c|}{\cellcolor[HTML]{FFD400}66.8} & \multicolumn{1}{c|}{} & \multicolumn{1}{c|}{\cellcolor[HTML]{FF0000}nan} & \multicolumn{1}{c|}{\cellcolor[HTML]{FF0000}2.5k} & \multicolumn{1}{c|}{\cellcolor[HTML]{609F00}248} & \multicolumn{1}{c|}{\cellcolor[HTML]{77FF00}41}  & \cellcolor[HTML]{64FF00}43  \\ \cline{2-13} 
\multicolumn{1}{|c|}{\multirow{-2}{*}{100$\times$}}  & \roastpp         & \multicolumn{1}{c|}{\cellcolor[HTML]{00FF00}75.2} & \multicolumn{1}{c|}{\cellcolor[HTML]{00FF00}75.8} & \multicolumn{1}{c|}{\cellcolor[HTML]{00FF00}75.2} & \multicolumn{1}{c|}{\cellcolor[HTML]{00FF00}74.9} & \multicolumn{1}{c|}{\cellcolor[HTML]{00FF00}75.1} & \multicolumn{1}{c|}{} & \multicolumn{1}{c|}{\cellcolor[HTML]{32FF00}48}  & \multicolumn{1}{c|}{\cellcolor[HTML]{37FF00}47}   & \multicolumn{1}{c|}{\cellcolor[HTML]{34FF00}48}  & \multicolumn{1}{c|}{\cellcolor[HTML]{33FF00}48}  & \cellcolor[HTML]{31FF00}48  \\ \hline
\multicolumn{1}{|c|}{}                               & \roast            & \multicolumn{1}{c|}{\cellcolor[HTML]{FF0000}10.0} & \multicolumn{1}{c|}{\cellcolor[HTML]{FF0000}10.0} & \multicolumn{1}{c|}{\cellcolor[HTML]{63FF00}53.1} & \multicolumn{1}{c|}{\cellcolor[HTML]{00FF00}56.0} & \multicolumn{1}{c|}{\cellcolor[HTML]{95FF00}52.1} & \multicolumn{1}{c|}{} & \multicolumn{1}{c|}{\cellcolor[HTML]{FF0000}nan} & \multicolumn{1}{c|}{\cellcolor[HTML]{FF0000}nan}  & \multicolumn{1}{c|}{\cellcolor[HTML]{F60900}549} & \multicolumn{1}{c|}{\cellcolor[HTML]{1FE000}116} & \cellcolor[HTML]{74FF00}41  \\ \cline{2-13} 
\multicolumn{1}{|c|}{\multirow{-2}{*}{1000$\times$}} & \roastpp         & \multicolumn{1}{c|}{\cellcolor[HTML]{72FF00}52.8} & \multicolumn{1}{c|}{\cellcolor[HTML]{8CFF00}52.3} & \multicolumn{1}{c|}{\cellcolor[HTML]{49FF00}53.6} & \multicolumn{1}{c|}{\cellcolor[HTML]{0BFF00}54.8} & \multicolumn{1}{c|}{\cellcolor[HTML]{95FF00}52.1} & \multicolumn{1}{c|}{} & \multicolumn{1}{c|}{\cellcolor[HTML]{50FF00}45}  & \multicolumn{1}{c|}{\cellcolor[HTML]{4DFF00}45}   & \multicolumn{1}{c|}{\cellcolor[HTML]{4AFF00}45}  & \multicolumn{1}{c|}{\cellcolor[HTML]{41FF00}46}  & \cellcolor[HTML]{4BFF00}45  \\ \hline
\multicolumn{1}{|c|}{}                               &                                   & \multicolumn{11}{c|}{\textbf{CIFAR-100 (VGG-11)}}                                                                                                                                                                                                                                                                                                                                                                                                                                                                                    \\ \hline
\multicolumn{1}{|c|}{}                               & \roast            & \multicolumn{1}{c|}{\cellcolor[HTML]{FF0000}1.0}  & \multicolumn{1}{c|}{\cellcolor[HTML]{21FF00}67.2} & \multicolumn{1}{c|}{\cellcolor[HTML]{15FF00}67.5} & \multicolumn{1}{c|}{\cellcolor[HTML]{E4FF00}63.6} & \multicolumn{1}{c|}{\cellcolor[HTML]{FF0000}19.8} & \multicolumn{1}{c|}{} & \multicolumn{1}{c|}{\cellcolor[HTML]{FF0000}8k}  & \multicolumn{1}{c|}{\cellcolor[HTML]{FF0000}659}  & \multicolumn{1}{c|}{\cellcolor[HTML]{00FF00}73}  & \multicolumn{1}{c|}{\cellcolor[HTML]{00FF00}67}  & \cellcolor[HTML]{00FF00}75  \\ \cline{2-13} 
\multicolumn{1}{|c|}{\multirow{-2}{*}{1.33$\times$}} & \roastpp         & \multicolumn{1}{c|}{\cellcolor[HTML]{00FF00}68.1} & \multicolumn{1}{c|}{\cellcolor[HTML]{00FF00}68.1} & \multicolumn{1}{c|}{\cellcolor[HTML]{00FF00}68.9} & \multicolumn{1}{c|}{\cellcolor[HTML]{00FF00}68.2} & \multicolumn{1}{c|}{\cellcolor[HTML]{00FF00}68.6} & \multicolumn{1}{c|}{} & \multicolumn{1}{c|}{\cellcolor[HTML]{0AF500}115} & \multicolumn{1}{c|}{\cellcolor[HTML]{0AF500}115}  & \multicolumn{1}{c|}{\cellcolor[HTML]{0AF500}115} & \multicolumn{1}{c|}{\cellcolor[HTML]{0AF500}115} & \cellcolor[HTML]{02FD00}109 \\ \hline
\multicolumn{1}{|c|}{}                               & \roast            & \multicolumn{1}{c|}{\cellcolor[HTML]{FF0000}38.7} & \multicolumn{1}{c|}{\cellcolor[HTML]{FF0000}1.0}  & \multicolumn{1}{c|}{\cellcolor[HTML]{00FF00}66.0} & \multicolumn{1}{c|}{\cellcolor[HTML]{FFE900}62.8} & \multicolumn{1}{c|}{\cellcolor[HTML]{FF0000}19.9} & \multicolumn{1}{c|}{} & \multicolumn{1}{c|}{\cellcolor[HTML]{FF0000}8k}  & \multicolumn{1}{c|}{\cellcolor[HTML]{00FF00}57}   & \multicolumn{1}{c|}{\cellcolor[HTML]{00FF00}106} & \multicolumn{1}{c|}{\cellcolor[HTML]{00FF00}68}  & \cellcolor[HTML]{00FF00}75  \\ \cline{2-13} 
\multicolumn{1}{|c|}{\multirow{-2}{*}{10$\times$}}   & \roastpp         & \multicolumn{1}{c|}{\cellcolor[HTML]{00FF00}65.8} & \multicolumn{1}{c|}{\cellcolor[HTML]{00FF00}66.5} & \multicolumn{1}{c|}{\cellcolor[HTML]{00FF00}66.1} & \multicolumn{1}{c|}{\cellcolor[HTML]{00FF00}66.0} & \multicolumn{1}{c|}{\cellcolor[HTML]{00FF00}66.3} & \multicolumn{1}{c|}{} & \multicolumn{1}{c|}{\cellcolor[HTML]{0DF200}117} & \multicolumn{1}{c|}{\cellcolor[HTML]{0DF200}117}  & \multicolumn{1}{c|}{\cellcolor[HTML]{0DF200}117} & \multicolumn{1}{c|}{\cellcolor[HTML]{0DF200}117} & \cellcolor[HTML]{0CF300}116 \\ \hline
\multicolumn{1}{|c|}{}                               & \roast            & \multicolumn{1}{c|}{\cellcolor[HTML]{FF0000}1.0}  & \multicolumn{1}{c|}{\cellcolor[HTML]{EAFF00}54.5} & \multicolumn{1}{c|}{\cellcolor[HTML]{0AFF00}57.1} & \multicolumn{1}{c|}{\cellcolor[HTML]{F9FF00}54.4} & \multicolumn{1}{c|}{\cellcolor[HTML]{FF0000}20.2} & \multicolumn{1}{c|}{} & \multicolumn{1}{c|}{\cellcolor[HTML]{FF0000}14k} & \multicolumn{1}{c|}{\cellcolor[HTML]{FF0000}1741} & \multicolumn{1}{c|}{\cellcolor[HTML]{798600}192} & \multicolumn{1}{c|}{\cellcolor[HTML]{00FF00}70}  & \cellcolor[HTML]{00FF00}75  \\ \cline{2-13} 
\multicolumn{1}{|c|}{\multirow{-2}{*}{100$\times$}}  & \roastpp         & \multicolumn{1}{c|}{\cellcolor[HTML]{00FF00}57.5} & \multicolumn{1}{c|}{\cellcolor[HTML]{00FF00}57.4} & \multicolumn{1}{c|}{\cellcolor[HTML]{00FF00}57.3} & \multicolumn{1}{c|}{\cellcolor[HTML]{00FF00}57.4} & \multicolumn{1}{c|}{\cellcolor[HTML]{00FF00}58.2} & \multicolumn{1}{c|}{} & \multicolumn{1}{c|}{\cellcolor[HTML]{00FF00}94}  & \multicolumn{1}{c|}{\cellcolor[HTML]{00FF00}94}   & \multicolumn{1}{c|}{\cellcolor[HTML]{00FF00}93}  & \multicolumn{1}{c|}{\cellcolor[HTML]{00FF00}94}  & \cellcolor[HTML]{00FF00}93  \\ \hline
\multicolumn{1}{|c|}{}                               & \roast            & \multicolumn{1}{c|}{\cellcolor[HTML]{FF0000}1.0}  & \multicolumn{1}{c|}{\cellcolor[HTML]{FFFB00}17.6} & \multicolumn{1}{c|}{\cellcolor[HTML]{00FF00}25.4} & \multicolumn{1}{c|}{\cellcolor[HTML]{10FF00}24.7} & \multicolumn{1}{c|}{\cellcolor[HTML]{FFEF00}17.1} & \multicolumn{1}{c|}{} & \multicolumn{1}{c|}{\cellcolor[HTML]{FF0000}nan} & \multicolumn{1}{c|}{\cellcolor[HTML]{FF0000}5595} & \multicolumn{1}{c|}{\cellcolor[HTML]{FF0000}675} & \multicolumn{1}{c|}{\cellcolor[HTML]{14EB00}122} & \cellcolor[HTML]{00FF00}72  \\ \cline{2-13} 
\multicolumn{1}{|c|}{\multirow{-2}{*}{1000$\times$}} & \roastpp         & \multicolumn{1}{c|}{\cellcolor[HTML]{39FF00}23.5} & \multicolumn{1}{c|}{\cellcolor[HTML]{3DFF00}23.4} & \multicolumn{1}{c|}{\cellcolor[HTML]{33FF00}23.7} & \multicolumn{1}{c|}{\cellcolor[HTML]{3AFF00}23.5} & \multicolumn{1}{c|}{\cellcolor[HTML]{36FF00}23.6} & \multicolumn{1}{c|}{} & \multicolumn{1}{c|}{\cellcolor[HTML]{00FF00}83}  & \multicolumn{1}{c|}{\cellcolor[HTML]{00FF00}85}   & \multicolumn{1}{c|}{\cellcolor[HTML]{00FF00}84}  & \multicolumn{1}{c|}{\cellcolor[HTML]{00FF00}84}  & \cellcolor[HTML]{00FF00}85  \\ \hline
\end{tabular}}
\end{table}
As we can see with {\roastpp} learning scheme, which we will describe in later part of the section, we can almost eliminate the sensitivity to the init-stdev and obtain the best results at all levels of init-stdev. Thus {\roastpp} reduces a hyper-parameter which turns out to be very sensitive towards the convergence of the RPS. 

\textbf{Why is global memory sharing in RPS unstable?:}\label{sec:why}
{\roast} performs global memory sharing, which means that parameters from the {\roast} array are shared across different components. {\roast} uses multiplicative scaling factors ( say $\lambda$s) to maintain the relative scales of different components. The $\lambda$s are  dependent on the init-sdev. For instance, if a component needs an initialization standard deviation of $0.1$ and the {\roast} array is initialized with init-stdev of $0.01$, then the multiplicative factor $\lambda$ for this component is $0.1/0.01 = 10$. These scale factors which have to be used due to global memory sharing are the root cause of this extreme sensitivity to init-stdev. we want to note that these issues are also present in HashedNet \citep{hashtrick}. But become more pronounced due to global memory sharing using multipliers.

This can be explained considering the effective change in value of parameters in a gradient descent(GD) step. Consider a parameter $x$ shared across $k$ different weights with scaling factor $\lambda_1,...,\lambda_k$. Then, we have the recovered weights $x_i = \lambda_i x$. The effective update of $x_i$ in one GD step can be written as follows where $x_i^{(t)}$ is recovered weight at time $t$ and $\mathcal{L}$ is the loss function.
\begin{equation}
    (x_i)^{(t+1)} - x_i^{(t)} = \lambda_i g(i) \frac{\partial \mathcal{L}}{\partial x } = \lambda_i g(i) \sum_{j=1}^k  \lambda_j g(j) \frac{\partial \mathcal{L}}{\partial x_j} \label{eq:effchange}
\end{equation}
These $\lambda$s can have varied values. For instance, the RESNET20 model has scales varying from $1$ to $5$ for init-sdev of $0.0588$. This implies that some parameters will see approximately $25\times$ larger effective updates than what they would see without parameter sharing. With larger effective gradients, it is unsurprising that the learning rates that worked for the full model will not work for the compressed model. Intuitively, with larger values of $\lambda$s, we will be forced to use smaller learning rates. We characterize this ``stability region" of learning rates, i.e., the range of learning rate $\eta$ that can be used with guaranteed convergence of gradient descent algorithm, in the next theorem.

\begin{theorem}(Stability of RPS with global memory sharing)\label{thm:opt1}
    Consider a function $F$, which is $L_f$ Lipschitz smooth, with $n$ parameters. Under RPS setup with RPS array of size $m$, hashing function $h$ and scaling factors $\lambda_1, \lambda_2 ,...,\lambda_n$. The stability range of gradient descent for learning rate goes from $(0, \frac{2}{L_f})$ for full model training to $\left(0, \frac{2}{\left( \max_{j=1}^m  \sum_{i=1, h(i) = j}^{i=n} \lambda_i^2  \right) L_f}\right)$ for compressed model 
\end{theorem}

The proof is presented in appendix \ref{app:convergence}. Thus, the possible range of learning rates that can be used for compressed models shrinks with increasing $\lambda$s (which depends on the choice of init-stdev) and increasing compression. 

\textbf{{\roastpp} gradient scaling scheme:}
We propose the following gradient scaling mechanism to remove the sensitivity to init-stdev and improve convergence. Before the update step, the gradients are scaled down using a scaler $\Gamma \in R^m$. So the update step becomes,
\begin{equation}
    \nabla_{\textrm{eff}} \mathcal{L} = \Gamma \circ \nabla \mathcal{L}
\end{equation}
where $\circ$ is element-wise multiplication. $\nabla_{\textrm{eff}}$ is used as the gradient in the subsequent optimizer steps. For instance in gradient descent the weights $x$ are updated as $x^{(t+1)} = x^{(t)} - \eta \nabla_{\textrm{eff}} \mathcal{L} $. We propose two gradient scalers. One directly follows from the above theorem \ref{thm:opt1}. The other is based on maintaining effective update sizes.

We define the \textbf{theory driven scaler} as follows,
\begin{equation}
    \Gamma[j] = \frac{1}{\sum_{i, h(i) = j} \lambda_i^2}
\end{equation}
The gradient of a parameter $j$ from the RPS array is scaled with a factor of $l_2$ norm of all component scaling factors used for recovery from this location $j$. With this new scaling, the modified stability range of RPS with this scheme essentially remains unchanged.
\begin{theorem}(Stability of RPS with gradient scaling)\label{thm:opt2}
      Consider a function F with $n$ parameters, which is Lipschitz smooth with constant $L_f$. Under RPS setup with RPS array of size $m$, hashing function $h$,  and scaling factors $\lambda_1 , \lambda_2 , ... \lambda_n$. The stability range of gradient descent for learning rate is maintained at $(0, \frac{2}{L_f})$ under compression when using the gradient scaler      $\Gamma[j] = \frac{1}{ \sum_{i, h(i) = j} \lambda_i^2}$
\end{theorem}
The proof is deferred to appendix \ref{app:convergence}. The gradient scaler effectively kills the effect of multiplicative factors. We also propose another way to scale below.

Another way of computing scaler is to maintain effective change in value of parameters after one gradient descent step. Under the same notation of parameter $x$ being shared across $x_1, ... x_k$ using scalers $\lambda_1, ... \lambda_k$, the effective change observed in a parameter $x_i$ after the update is given in equation \ref{eq:effchange}.
If we use scaling $\gamma$, then,
\begin{equation}
\textrm{average-effective-update} =  \frac{1}{k} \gamma \Big(\sum_{i=1}^k \lambda_i\Big) {\left(\sum_{i=1}^k  \lambda_i \frac{\partial \mathcal{L}}{\partial x_i}\right)}
\end{equation}
We want to scale the change with $\gamma$ such that the effective change of all the recovered weights is similar to those with and without parameter sharing.
\begin{equation}
\underbracket[,6pt]{\frac{1}{k}  \sum_{i=1}^k  \frac{\partial \mathcal{L}}{ \partial x_i}  }_{\textrm{simple average}} \quad [\textrm{full}] \qquad \approx \qquad \frac{1}{k} \gamma \Big(\sum_{i=1}^k \lambda_i\Big)^2 \underbracket[,6pt]{\frac{\left(\sum_{i=1}^k  \lambda_i \frac{\partial \mathcal{L}}{\partial x_i}\right)}{\sum_{i=1}^k \lambda_k}}_{\textrm{weighted average}} \quad [\textrm{RPS}]
\end{equation}

We propose $\gamma = \frac{k}{ \left(\sum_{i=1}^k \lambda_i\right)^2}$ and thus, the \textbf{effective update scaler} for each parameter in RPS is,
\begin{equation}
    \Gamma[j] = \frac{\sum_{i=1, h(i) = j} ^{i=n}1} {(\sum_{i=1, h(i) = j}^{i=n} \lambda_i)^2} \forall j \in \{0,...m-1\}
\end{equation}
While both scaling mechanisms work well in stabilizing the convergence and make it robust to init-stdev, we use  effective average-based gradient scaler throughout the paper including Table~\ref{fig:stability} \footnote{Our initial implementation of theoretical-scaler had a bug which prompted us to use effective-update-scaler in our experiments. When fixed, theoretical-scaler performs better than effective-update-scaler and we present results in appendix \ref{app:scaling_comparison}. This difference is inconsequential to our overall argument}. 
%We provide additional results comparing the two gradient scaling mechanisms in appendix \ref{app:convergence}.
\ssection{Pareto-Continuity of {\roastpp}}\label{sec:pareto}
\begin{figure}
    \centering
    \includegraphics[scale=0.24]{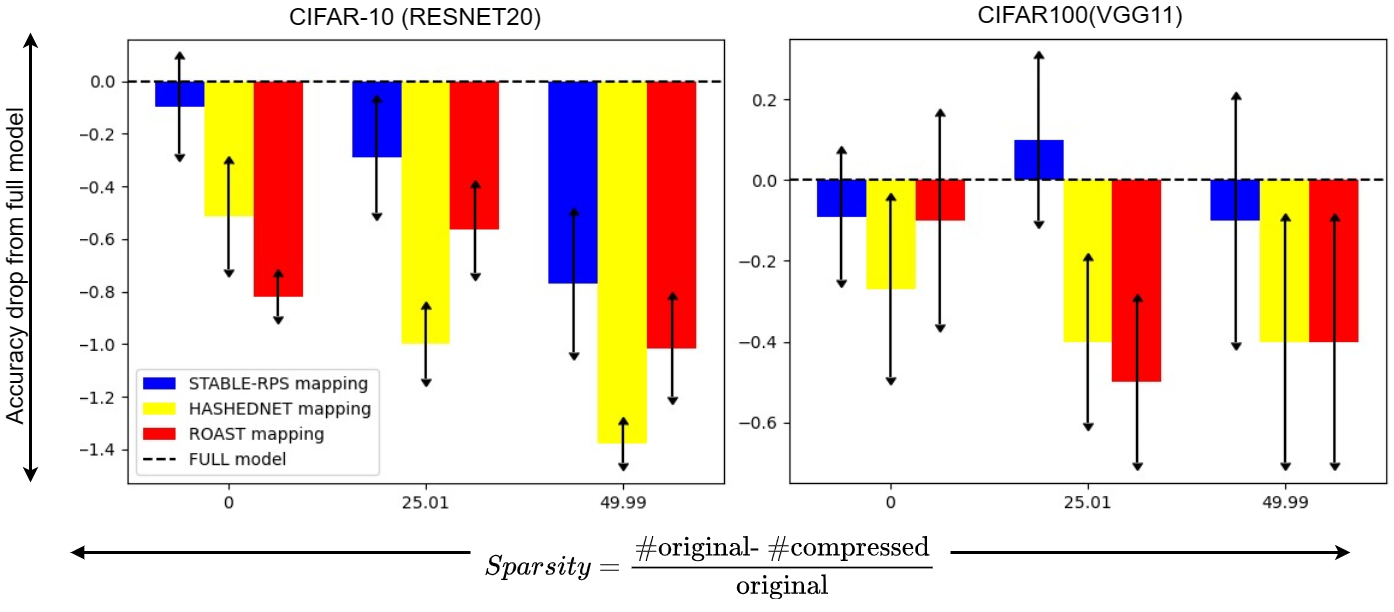}
    \caption{Pareto-continuity of {\roastpp}, {\roast} and {HashedNet} mapping (all mappings are used with effective update scaler described in section \ref{sec:stability}). {\roastpp} shows Pareto-continuity and improvement over other mappings.} 
     \vspace{-0.1cm}
    \label{fig:pareto}
    \label{fig:mapping}
    %\vspace{-0.5cm}
\end{figure}
This section discusses the mappings used in RPS and their shortcomings concerning Pareto-continuity. Early papers on RPS used element-wise mapping where each weight is independently mapped into the weight repository. Later this idea was extended to mapping chunks to improve cache-efficiency. All these mappings lack one fundamental property that all compression techniques should have, namely \textit{pareto-continuity} - as memory budget tends to full model memory, the mapping should revert to one-to-one mapping, causing no-collisions and thus recover exact full model (modulo permuted weight layout in memory). Pruning-based compression trivially has this property. The way randomized hashing is used in these methods, there will be some collisions even at full memory. A related issue with these mappings is that they do not have optimal \textit{load factor}. The load factor is the maximum number of collisions under a hash function. If a hash function maps from $[n]$ to $[m]$ where $[n] = \{0,...,n-1\}$, then optimal \textit{load factor}  is $\lceil n / m \rceil $. However, for these mappings, the \textit{load factor} is usually much higher than optimal. A mapping that has optimal load factor will have Pareto-continuity property as well.

\textbf{{\roastpp} mapping:} We want to build a mapping that (1) has optimal load factor ( and hence, Pareto-continuity) and (2) maintains the cache-efficiency of {\roast}. An illustration of {\roastpp} mapping is shown in the figure \ref{fig:expmapping}. Consider {\roast} mapping. It divides the model into different-shaped chunks. We flatten out the chunks into a single array of weights. The location of each weight, say $i$, in this array is called its global index $\mathcal{G}(i)$. We perform what we call {\randomfold} mapping on the global index.

A {\randomfold} hash function $h_u : [n] \rightarrow [m]$ where $[n] = \{0,...,n-1\}$ which uses a random hash function $u$  is defined as,
\begin{equation}
     h_u(x) = (u( \lfloor x/m \rfloor ) + (x \% m)) \% m
\end{equation}
The {\randomfold} mapping divides the range  $[n]$ into partitions of size $m$ and then applies a random hash function on each partition number. The mapping of a partition, then, is circular and wraps around the $m$ (see figure \ref{fig:expmapping}). The {\roastpp} mapping is then defined as
\begin{equation}
    \textrm{\roastpp}(i) = h_u(\mathcal{G}(i))
\end{equation}
We summarize the properties of {\roastpp} in the theorem below,
\begin{theorem}
    The {\roastpp} hash function has 
    \begin{enumerate}[nosep, leftmargin=*]
        \item optimal load property and, thus, Pareto-continuity
        \item Under the assumption that {\roast} chunks are smaller than memory budget, the number of cache-line fetches under {\roastpp} for each {\roast} chunk is $\mathcal{C}({\textrm{\roastpp}}) \leq  \mathcal{C}({\textrm{\roast}}) + 3$ where $\mathcal{C}$ denotes number of cache-line fetches for a {\roast} chunk.
    \end{enumerate}
\end{theorem}
The proof is presented in appendix \ref{app:mapping}. The {\roastpp} has an optimal load factor, which leads to pareto-continuity. It pays a slight cost regarding cache efficiency due to {\randomfold} potentially separating {\roast} chunks into different partitions. 

In our empirical evaluation, we try the abovementioned three mappings - element-wise mapping, {\roast} mapping, and {\roastpp} mapping. We use all these mappings with gradient scaling improvements of {\roastpp} for stable training. The results are shown in figure \ref{fig:pareto}. As we can see, element-wise mapping and {\roast} mapping do not achieve full model accuracy at full model memory. This is overcome by {\roastpp} mapping. Also, {\roastpp} mapping outperforms the other two mappings at higher compression rates as well due to its optimal load property, which minimizes the maximum collisions per bucket.
\ssection{Parameter Sharing vs. Other options}
\begin{figure}
\centering
\includegraphics[scale=0.2]{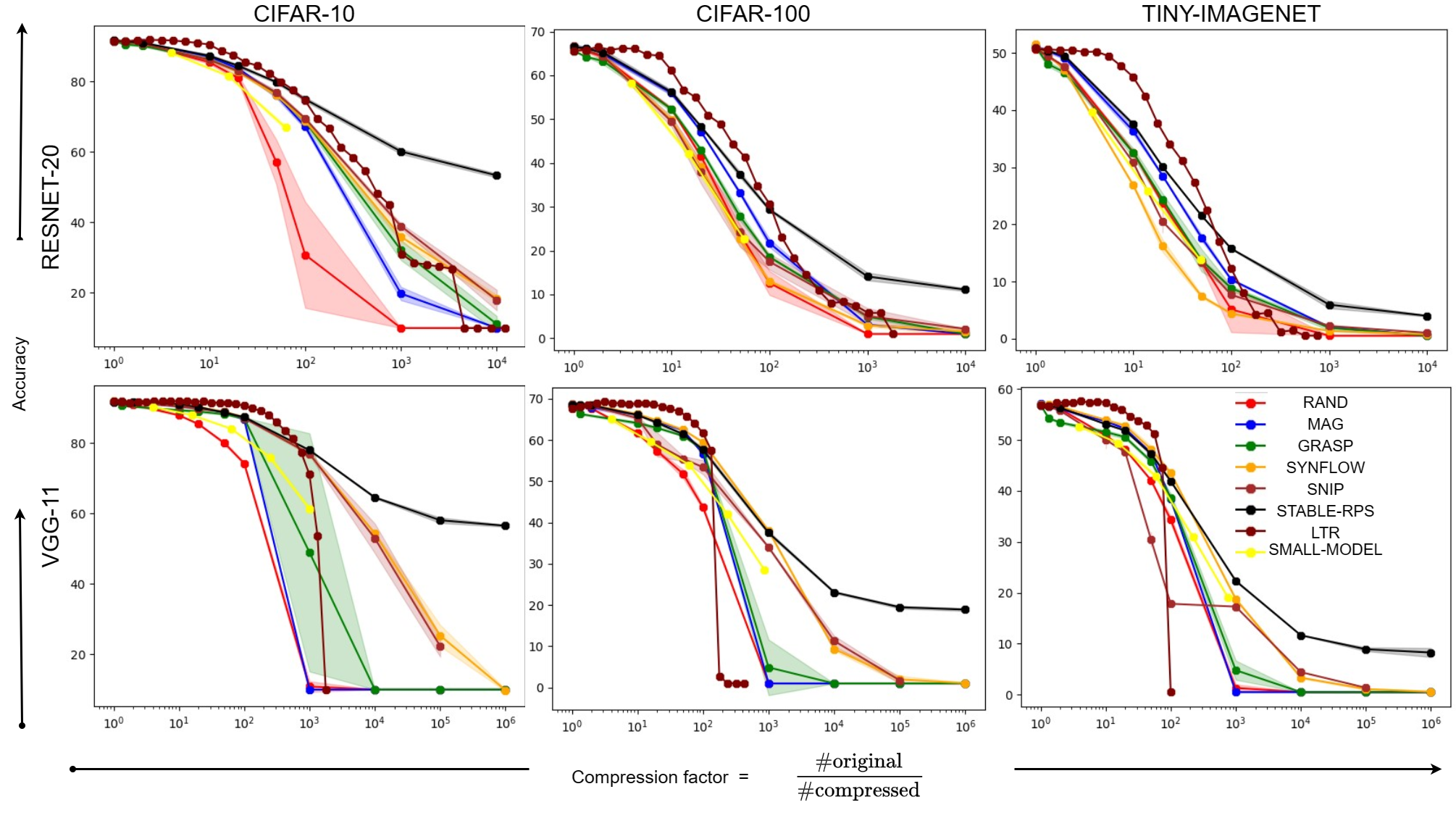}
\caption{Memory-Accuracy tradeoff of {\roastpp} against different pruning methods and small models for training. This figure was generated using approximately 1300 experiments. All of this tradeoff-data will be made public so that researchers can use it for plotting baselines.}
    \label{fig:1}
    %\vspace{-0.5cm}
\end{figure}
\subsection{Empirical study}
We first empirically demonstrate the advantage of {\roastpp} parameter sharing method on vision tasks. We extensively evaluate memory-accuracy tradeoffs on various model reduction techniques. The results presented in this section summarize over 1300+ experiments performed on V100 and Quadro RTX 8000 GPUs. We reestablish some known results on pruning, stress test existing results on higher and unexplored orders of compression, and add more methods to the mix. We compare the following methods.
\begin{itemize}[leftmargin=*, nosep]
\item {\roastpp} (uninformed): Model, data, and initialization agnostic parameter sharing method.

\item {\rand} (uninformed): Model, data, and initialization agnostic pruning technique. In terms of information, this is equal to \roastpp.

\item {\mg}, {\snip}, {\synflow}, {\grasp} (Moderately informed). These techniques use data, initialization of model, or both. As pointed out by \cite{tanaka2020pruning}, most of these methods ( except for {\grasp} ) benefit from iterative pruning instead of one-shot pruning. We use $100$ pruning iterations for all of them except {\grasp}, for which we perform one-shot pruning. 
\item Lottery ticket Rewinding({\ltr}) (Highly informed) We rewind the weights to $500$ iterations for all the experiments in line with the observations made by \citet{pmlr-v119-frankle20a}. While there has been much research on finding winning tickets early, we use {\ltr} as a representative method for the memory-accuracy tradeoff that is achievable using these methods.

\item {\smallmodel} (uninformed): To reduce the model size while using standard modules, we reduce the hidden dimensions of the model while keeping the depth same.
\end{itemize}

In Figure \ref{fig:1}, on the x-axis we have compression factor = $\frac{\mathrm{original}}{\mathrm{compressed}}$. In all of the methods and models we do not compress bias and batch-norm parameters. So the ratio is computed over remaining parameters. A factor of $10$ means only 10\% of parameters remain. We show error bars, which are $\pm$ standard deviations of accuracy. Most experimental settings (except {\ltr}) are run on $3$ or more seeds. The exact experimental details (learning rate schedules, hyper-parameters,etc.) for pruning, {\roastpp} and {\ltr} are provided in appendix \ref{sec:app_experimentdetails}.
We make the following observations,

\textbf{{\roastpp} vs. Others}
\begin{itemize}[nosep, leftmargin=*]
\item In a comparison of uninformed choices, {\roastpp} outperforms {\rand} at all levels of compression. This shows that under similar information, the capacity of {\roastpp} is strictly better than {\rand} pruning. We will revisit this comparison theoretically towards the end of the section.

\item When compared with moderately informed pruning techniques, {\roastpp}, which is uninformed, is always superior in higher compression regions and competitive or better in the lower-compression areas. A highly informed pruning method such as LTR is also worse in high compression regimes when compared to an uninformed {\roastpp}. The fact that even the most informed pruning cannot beat uninformed parameter sharing at high compression highlights the severity of capacity-disadvantage pruning as a paradigm faces in comparison with {\roastpp}.
\item  {\smallmodel} performance is significantly inferior to {\roastpp} 
\end{itemize}
\textbf{Moderately informed Pruning and {\ltr}}
\begin{itemize}[nosep, leftmargin=*]
\item There is no one clear winner in moderately informed pruning methods. This does not contradict results presented by \citet{tanaka2020pruning} since we implement iterative versions of pruners, which improve over their baseline single-shot performance. 
\item We evaluate {\ltr} at higher compression, which was not done before. We find that {\ltr} experiences layer collapse, making it, at times, worse than moderately informed pruning methods. This confirms the discussion by \citet{tanaka2020pruning}, which shows that {\ltr} balances the weight magnitudes with expensive training, but this balancing is imperfect. It implies that {\ltr} can be improved in high compression regions potentially using ideas from \citet{tanaka2020pruning}.
\end{itemize}
\textbf{{\smallmodel} vs. Others}.
\begin{itemize}[nosep, leftmargin=*]
\item {\smallmodel} generally gives a worse tradeoff than most moderately informed pruning and {\ltr}. However, its tradeoff curve is better than {\rand}.
This implies that a bit informed sparse networks are more powerful than constructing \smallmodel with standard modules.
\item The difference in the capacity of a small model and {\roastpp} is significantly large. Also, in this case, {\roastpp} is not data/initialization informed (unlike pruning). It shows that there is a potential for us to discover new standard modules based on parameter sharing that can be better than standard modules such as Linear and Convolutions.
\end{itemize}

\ssubsection{Pruning vs. {\roastpp} : theoretical view}
We now analyze the effectiveness of {\rand} pruning and {\roastpp} in a theoretical setting. We first analyze the dimensionality reduction problem (preserving norms and inner products under data compression). In data compression, pruning implies dropping vector components, and {\roastpp} means sketching vector components under the specific mapping function. We later show that the quality of learning linear models can be reduced to norm preservation under data compression.
One important thing to note here is that we get these theoretical results only for {\roastpp} mapping, and these results would not have been possible if we were analyzing other previously used mappings.

\textbf{Data compression:} Consider two vectors $x,y \in R^n$. Let the parameter budget be $m = n/k $. For the sake of simplicity, we assume that $k$ is an integer. Let $\hat{x}, \hat{y}$ be the compressed $x$ and $y$. The estimation of inner products is,
\begin{equation}
    \langle \hat{x}, \hat{y} \rangle_{\textrm{prune}} = k \sum_{j=1}^n x_j y_j \mathbf{1}(j)
\end{equation}
\vspace{-0.6cm}
\begin{equation}
    \langle \hat{x}, \hat{y} \rangle_{\textrm{\roastppsmall}} = \sum_{i=1}^m \Big( \sum_{j=1}^n x_j g(j) \mathbf{1}(h(j) = i) \Big) \Big( \sum_{j=1}^n y_j g(j) \mathbf{1}(h(j) = i) \Big)
    \vspace{-0.25cm}
\end{equation}
where $\mathbf{1}(i)$ is an indicator if $i^{th}$ component is sampled under pruning and $\mathbf{1}(h(j) = i)$ is an indicator if the hash mapping $h$ maps $j$ to $i$ and $g(j)$ is the sign hash function. The following theorem shows the theoretical advantage of {\roastpp} for preserving norms. We have a detailed analysis of inner products and norms in appendix \ref{app:theory}. 

\begin{theorem}
    Given vectors $x\in R^n$ under pruning and {\roastpp} with memory budget $m = n / k$ for $k \in \mathbf{N}$, the estimates for both methods are unbiased. The variances are as follows,
    \begin{equation}
    \mathbf{V}_{\textrm{sample}}(\langle \hat{x}, \hat{x} \rangle_{prune}) =   \Big( \frac{n}{m} - 1 \Big) \sum_{i=1}^n  x_i^4  + \left( \frac{(m-n)}{m(n-1)}\right)  \sum_{i\neq j; 
 i,j=1}^n x_i^2 x_j^2 
    \end{equation}
    \begin{align}
    \mathbf{V}_{h}(\langle \hat{x}, \hat{x} \rangle_{\roastppsmall}) = \frac{2}{m} \frac{n-m}{n-1} \Big( \sum_{i \neq j} x_i^2 x_j^2 \Big)
\end{align}
Let $x_i$ and $y_i$ be i.i.d drawn from a distribution with $\mathcal{N}(\mu=0, \sigma=\sigma_i)$. Let this data distribution be $\mathcal{D}$. Then we have,
\begin{equation}
 \mathbf{E}_\mathcal{D}(\mathbf{V}_{sample}(\langle \hat{x}, \hat{y} \rangle_{
 \textrm{prune}
 })) \geq \mathbf{E}_\mathcal{D}(\mathbf{V}_{h}(\langle \hat{x}, \hat{y} \rangle)_{\roastppsmall}) 
\end{equation}
with equality being achieved if and only if all $\sigma_i$s are equal.
\end{theorem}
\textbf{Interpretation:} It is clear that the variance of approximation methods depends on what $x$ we are considering. So, we want to understand for what portion of data is pruning or {\roastpp} superior. Hence, we take the expectation over data. In the case of a perfectly balanced distribution (which implies a radially symmetric distribution), the expectations are equal. So exactly the same proportions of data points favor pruning  and {\roastpp}. However, in practice, distribution is hardly symmetric. In fact, distributions are generally power-law; in this situation, {\roastpp} is strictly better in terms of portions of data-space that it approximates better. The actual advantage will depend on the scale of power-law with a higher power-law favoring {\roastpp}.

\textbf{Model Compression:} Now let us look at pruning and {\roastpp} in the context of model parameters for linear models. Consider the learning of a linear model for $y$ given signal vector $x_1, ..., x_n$ with independent components. Let the correlation of output variable $y$ with each $x_i$ be $\rho_i$. Let the vector $\rho = (\rho_1, \rho_2, ... \rho_n)$. We can then analyze the best residual obtained (i.e., learned model residual) once the pruning sample of weights or mapping of {\roastpp} has been decided. The following result reduces the residual obtained under model compression to data compression on the vector $\rho$.

\begin{theorem}
    The optimal (least) residual for linear regression problem for target $y$ with standard deviation $\sigma_y$ and independent input signals $(x_1, x_2, ..., x_n)$ each of which has standard deviation $\sigma_x$, with correlations $\rho_i = \frac{Cov(y, x_i)}{\sigma_x \sigma_y}$, can be written as,
\begin{equation}
    \frac{\textrm{Res}_{comp}(\hat{y}) - \textrm{Res}(\hat{y})}{\textrm{Res}(\hat{y})} =  \frac{\langle \rho , \rho \rangle - \frac{1}{k} \langle \hat{\rho}, \hat{\rho} \rangle }{1 - \langle \rho , \rho \rangle}
\end{equation}
    where $\rho$ is the vector of all $\rho_i$s and $\langle \hat{\rho}, \hat{\rho} \rangle $ is the estimate of norm of $\rho$ vector under data compression (pruning or {\roastpp}) and $\textrm{Res}_{comp}$ is residual under compression.
\end{theorem}

\textbf{Interpretation:} The residual obtained by the compressed model (both pruning and {\roastpp}) depends on how well the norm of the vector $\rho$ is preserved under data compression. As we have seen in the previous theorem, if the correlations are not balanced (which is usually the case in actual data), then {\roastpp} is better than pruning on average.
\vspace{-0.2cm}
\ssection{Discussion and Conclusion}
Pruning has been one of the most successful tools for model compression - post-training or at the start of training. This paper brings to light the capacity issue with sparse models obtained via pruning, which limits the success of even the most intelligent pruning schemes. On the other hand, parameter sharing is much more powerful. Uninformed parameter sharing is already better than the best pruning scheme at high compression and competitive or better than moderately informed pruning at all compression. This paper argues in favor of a paradigm shift toward parameter-sharing-based compression. 

Many directions need exploring to fully realize the power of parameter sharing. We briefly discuss them here. This paper talks about parameter sharing from the start of training and shows the benefits of the parameter-sharing paradigm. To widely realize the benefits of parameter sharing, we need to be able to develop a post-training parameter-sharing scheme. Another issue with parameter sharing in its current form is that it requires the same computation as the original architecture. To make it practical, we must develop implementations leveraging parameter sharing to reduce computational load. Additionally, an interesting question is whether we can leverage the power of parameter sharing to design some fundamental layer of neural network - which is low memory, low computation but high capacity. Finally, the most immediate progress in this research direction can be made if we devise informed parameter sharing strategies.

\clearpage
\bibliography{refs-ssm}

\begin{thebibliography}{27}
\providecommand{\natexlab}[1]{#1}
\providecommand{\url}[1]{\texttt{#1}}
\expandafter\ifx\csname urlstyle\endcsname\relax
  \providecommand{\doi}[1]{doi: #1}\else
  \providecommand{\doi}{doi: \begingroup \urlstyle{rm}\Url}\fi

\bibitem[Alizadeh et~al.(2022)Alizadeh, Tailor, Zintgraf, van Amersfoort,
  Farquhar, Lane, and Gal]{prospr}
Milad Alizadeh, Shyam~A Tailor, Luisa~M Zintgraf, Joost van Amersfoort,
  Sebastian Farquhar, Nicholas~Donald Lane, and Yarin Gal.
\newblock Prospect pruning: Finding trainable weights at initialization using
  meta-gradients.
\newblock \emph{arXiv preprint arXiv:2202.08132}, 2022.

\bibitem[Bellec et~al.(2017)Bellec, Kappel, Maass, and
  Legenstein]{bellec2017deep}
Guillaume Bellec, David Kappel, Wolfgang Maass, and Robert Legenstein.
\newblock Deep rewiring: Training very sparse deep networks.
\newblock \emph{arXiv preprint arXiv:1711.05136}, 2017.

\bibitem[Chen et~al.(2020)Chen, Medini, Farwell, Tai, Shrivastava,
  et~al.]{chen2020slide}
Beidi Chen, Tharun Medini, James Farwell, Charlie Tai, Anshumali Shrivastava,
  et~al.
\newblock Slide: In defense of smart algorithms over hardware acceleration for
  large-scale deep learning systems.
\newblock \emph{Proceedings of Machine Learning and Systems}, 2:\penalty0
  291--306, 2020.

\bibitem[Chen et~al.(2015)Chen, Wilson, Tyree, Weinberger, and Chen]{hashtrick}
Wenlin Chen, James Wilson, Stephen Tyree, Kilian Weinberger, and Yixin Chen.
\newblock Compressing neural networks with the hashing trick.
\newblock In \emph{International conference on machine learning}, pp.\
  2285--2294. PMLR, 2015.

\bibitem[Desai \& Shrivastava(2022)Desai and Shrivastava]{NEURIPS2022_dbae9151}
Aditya Desai and Anshumali Shrivastava.
\newblock The trade-offs of model size in large recommendation models : 100gb
  to 10mb criteo-tb dlrm model.
\newblock In S.~Koyejo, S.~Mohamed, A.~Agarwal, D.~Belgrave, K.~Cho, and A.~Oh
  (eds.), \emph{Advances in Neural Information Processing Systems}, volume~35,
  pp.\  33961--33972. Curran Associates, Inc., 2022.
\newblock URL
  \url{https://proceedings.neurips.cc/paper_files/paper/2022/file/dbae915128892556134f1c5375855590-Paper-Conference.pdf}.

\bibitem[Desai et~al.(2021)Desai, Pan, Sun, Chou, and
  Shrivastava]{desai2021semantically}
Aditya Desai, Yanzhou Pan, Kuangyuan Sun, Li~Chou, and Anshumali Shrivastava.
\newblock Semantically constrained memory allocation (scma) for embedding in
  efficient recommendation systems.
\newblock \emph{arXiv preprint arXiv:2103.06124}, 2021.

\bibitem[Desai et~al.(2022)Desai, Chou, and Shrivastava]{MLSYS2022_1eb34d66}
Aditya Desai, Li~Chou, and Anshumali Shrivastava.
\newblock Random offset block embedding (robe) for compressed embedding tables
  in deep learning recommendation systems.
\newblock In D.~Marculescu, Y.~Chi, and C.~Wu (eds.), \emph{Proceedings of
  Machine Learning and Systems}, volume~4, pp.\  762--778, 2022.
\newblock URL
  \url{https://proceedings.mlsys.org/paper_files/paper/2022/file/1eb34d662b67a14e3511d0dfd78669be-Paper.pdf}.

\bibitem[Desai et~al.(2023)Desai, Zhou, and Shrivastava]{desai2023hardware}
Aditya Desai, Keren Zhou, and Anshumali Shrivastava.
\newblock Hardware-aware compression with random operation access specific tile
  (roast) hashing.
\newblock In \emph{International Conference on Machine Learning}, pp.\
  7732--7749. PMLR, 2023.

\bibitem[Evci et~al.(2020)Evci, Gale, Menick, Castro, and
  Elsen]{evci2020rigging}
Utku Evci, Trevor Gale, Jacob Menick, Pablo~Samuel Castro, and Erich Elsen.
\newblock Rigging the lottery: Making all tickets winners.
\newblock In \emph{International Conference on Machine Learning}, pp.\
  2943--2952. PMLR, 2020.

\bibitem[Frankle \& Carbin(2018)Frankle and Carbin]{frankle2018lottery}
Jonathan Frankle and Michael Carbin.
\newblock The lottery ticket hypothesis: Finding sparse, trainable neural
  networks.
\newblock \emph{arXiv preprint arXiv:1803.03635}, 2018.

\bibitem[Frankle et~al.(2020{\natexlab{a}})Frankle, Dziugaite, Roy, and
  Carbin]{pmlr-v119-frankle20a}
Jonathan Frankle, Gintare~Karolina Dziugaite, Daniel Roy, and Michael Carbin.
\newblock Linear mode connectivity and the lottery ticket hypothesis.
\newblock In Hal~Daumé III and Aarti Singh (eds.), \emph{Proceedings of the
  37th International Conference on Machine Learning}, volume 119 of
  \emph{Proceedings of Machine Learning Research}, pp.\  3259--3269. PMLR,
  13--18 Jul 2020{\natexlab{a}}.
\newblock URL \url{https://proceedings.mlr.press/v119/frankle20a.html}.

\bibitem[Frankle et~al.(2020{\natexlab{b}})Frankle, Dziugaite, Roy, and
  Carbin]{frankle2020pruning}
Jonathan Frankle, Gintare~Karolina Dziugaite, Daniel~M Roy, and Michael Carbin.
\newblock Pruning neural networks at initialization: Why are we missing the
  mark?
\newblock \emph{arXiv preprint arXiv:2009.08576}, 2020{\natexlab{b}}.

\bibitem[Han et~al.(2015)Han, Pool, Tran, and Dally]{han2015learning}
Song Han, Jeff Pool, John Tran, and William Dally.
\newblock Learning both weights and connections for efficient neural network.
\newblock \emph{Advances in neural information processing systems}, 28, 2015.

\bibitem[Hassibi \& Stork(1992)Hassibi and Stork]{hassibi1992second}
Babak Hassibi and David Stork.
\newblock Second order derivatives for network pruning: Optimal brain surgeon.
\newblock \emph{Advances in neural information processing systems}, 5, 1992.

\bibitem[Howard et~al.(2017)Howard, Zhu, Chen, Kalenichenko, Wang, Weyand,
  Andreetto, and Adam]{howard2017mobilenets}
Andrew~G Howard, Menglong Zhu, Bo~Chen, Dmitry Kalenichenko, Weijun Wang,
  Tobias Weyand, Marco Andreetto, and Hartwig Adam.
\newblock Mobilenets: Efficient convolutional neural networks for mobile vision
  applications.
\newblock \emph{arXiv preprint arXiv:1704.04861}, 2017.

\bibitem[Iandola et~al.(2016)Iandola, Han, Moskewicz, Ashraf, Dally, and
  Keutzer]{iandola2016squeezenet}
Forrest~N Iandola, Song Han, Matthew~W Moskewicz, Khalid Ashraf, William~J
  Dally, and Kurt Keutzer.
\newblock Squeezenet: Alexnet-level accuracy with 50x fewer parameters and< 0.5
  mb model size.
\newblock \emph{arXiv preprint arXiv:1602.07360}, 2016.

\bibitem[Jaderberg et~al.(2014)Jaderberg, Vedaldi, and
  Zisserman]{jaderberg2014speeding}
Max Jaderberg, Andrea Vedaldi, and Andrew Zisserman.
\newblock Speeding up convolutional neural networks with low rank expansions.
\newblock \emph{arXiv preprint arXiv:1405.3866}, 2014.

\bibitem[LeCun et~al.(1990)LeCun, Boser, Denker, Henderson, Howard, Hubbard,
  Jackel, and Touretzky]{lecun1990advances}
Yann LeCun, B~Boser, John~S Denker, Donnie Henderson, RE~Howard, Wayne~E
  Hubbard, LD~Jackel, and DS~Touretzky.
\newblock Advances in neural information processing systems.
\newblock \emph{San Francisco, CA, USA: Morgan Kaufmann Publishers Inc}, pp.\
  396--404, 1990.

\bibitem[Lee et~al.(2018)Lee, Ajanthan, and Torr]{lee2018snip}
Namhoon Lee, Thalaiyasingam Ajanthan, and Philip~HS Torr.
\newblock Snip: Single-shot network pruning based on connection sensitivity.
\newblock \emph{arXiv preprint arXiv:1810.02340}, 2018.

\bibitem[Mocanu et~al.(2018)Mocanu, Mocanu, Stone, Nguyen, Gibescu, and
  Liotta]{mocanu2018scalable}
Decebal~Constantin Mocanu, Elena Mocanu, Peter Stone, Phuong~H Nguyen,
  Madeleine Gibescu, and Antonio Liotta.
\newblock Scalable training of artificial neural networks with adaptive sparse
  connectivity inspired by network science.
\newblock \emph{Nature communications}, 9\penalty0 (1):\penalty0 2383, 2018.

\bibitem[Mozer \& Smolensky(1988)Mozer and Smolensky]{mozer1988skeletonization}
Michael~C Mozer and Paul Smolensky.
\newblock Skeletonization: A technique for trimming the fat from a network via
  relevance assessment.
\newblock \emph{Advances in neural information processing systems}, 1, 1988.

\bibitem[Novikov et~al.(2015)Novikov, Podoprikhin, Osokin, and
  Vetrov]{novikov2015tensorizing}
Alexander Novikov, Dmitrii Podoprikhin, Anton Osokin, and Dmitry~P Vetrov.
\newblock Tensorizing neural networks.
\newblock \emph{Advances in neural information processing systems}, 28, 2015.

\bibitem[Pham et~al.(2018)Pham, Guan, Zoph, Le, and Dean]{Pham2018_ilh}
Hieu Pham, Melody Guan, Barret Zoph, Quoc Le, and Jeff Dean.
\newblock Efficient neural architecture search via parameters sharing.
\newblock In Jennifer Dy and Andreas Krause (eds.), \emph{Proceedings of the
  35th International Conference on Machine Learning}, volume~80 of
  \emph{Proceedings of Machine Learning Research}, pp.\  4095--4104. PMLR,
  10--15 Jul 2018.
\newblock URL \url{https://proceedings.mlr.press/v80/pham18a.html}.

\bibitem[Prabhu et~al.(2018)Prabhu, Varma, and Namboodiri]{prabhu2018deep}
Ameya Prabhu, Girish Varma, and Anoop Namboodiri.
\newblock Deep expander networks: Efficient deep networks from graph theory.
\newblock In \emph{Proceedings of the European Conference on Computer Vision
  (ECCV)}, pp.\  20--35, 2018.

\bibitem[Spring \& Shrivastava(2017)Spring and Shrivastava]{spring2017scalable}
Ryan Spring and Anshumali Shrivastava.
\newblock Scalable and sustainable deep learning via randomized hashing.
\newblock In \emph{Proceedings of the 23rd ACM SIGKDD International Conference
  on Knowledge Discovery and Data Mining}, pp.\  445--454, 2017.

\bibitem[Tanaka et~al.(2020)Tanaka, Kunin, Yamins, and
  Ganguli]{tanaka2020pruning}
Hidenori Tanaka, Daniel Kunin, Daniel~L Yamins, and Surya Ganguli.
\newblock Pruning neural networks without any data by iteratively conserving
  synaptic flow.
\newblock \emph{Advances in neural information processing systems},
  33:\penalty0 6377--6389, 2020.

\bibitem[Zhang et~al.(2022)Zhang, Wang, and He]{zhang2022grasp}
Minjia Zhang, Wenhan Wang, and Yuxiong He.
\newblock Grasp: Optimizing graph-based nearest neighbor search with subgraph
  sampling and pruning.
\newblock In \emph{Proceedings of the Fifteenth ACM International Conference on
  Web Search and Data Mining}, pp.\  1395--1405, 2022.

\end{thebibliography}
\bibliographystyle{iclr2024_conference}

\appendix
\section{Code details}
The code can be found here :
\begin{itemize}
    \item RPS implementations : https://github.com/apd10/FakeRoast 
    \item Experiments for pruning and RPS : We modify original Synaptic flow repository here and add new experiments to it : https://github.com/apd10/Synaptic-Flow 
    \item LTR is run using https://github.com/facebookresearch/open\_lth
\end{itemize}

\clearpage
\section{Theory : Dimensionality reduction analysis on Inner-Products and norms} \label{app:theory}

Consider two vectors $x$ and $y$ in $R^n$ space. There are two methods to perform dimensoinality reduction 

\begin{enumerate}
    \item Subsample the vector ( making it sparse / pruning ) The transformation is 
    \begin{equation}
        \hat{x} = \sqrt{\frac{n}{m}} Sx 
    \end{equation}
    where S is a $m \times n$ sampling matrix composed of only zeros and ones. Also, each row has exactly one element non-zero. Each column has a maximum of one non-zero.
    \item Projection ( randomly combining the different elements ) In this case, we perform the following transformation
    \begin{equation}
        \hat{x} = R P x
    \end{equation}
    where P is a random permutation matrix and R is a $m \times n$ matrix. R is the random {\roastpp} matrix.
\end{enumerate}

Let us look at the inner product preservation in both cases. 
\subsection{pruning}
the estimator of inner products
\begin{equation}
    \langle \hat{x}, \hat{y} \rangle = \frac{n}{m} \sum_{i=1}^n x_i y_i \mathbf{1}(i)
\end{equation}
where $\mathbf{1}(i)$ implies that $i$ is selected.

\begin{equation}
    \mathbf{E}(\langle \hat{x}, \hat{y} \rangle) = \frac{n}{m} \frac{m}{n} \sum_{i=1}^n x_i y_i = \langle x, y \rangle 
\end{equation}

\begin{align}
    (\langle \hat{x}, \hat{y} \rangle)^2 = &  \frac{n^2}{m^2} \left( \sum_{i=1}^n x_i y_i \mathbf{1}(i) \right)^2 \\
    =  & \frac{n^2}{m^2} \left( \sum_{i,j=1}^n x_i y_i x_j y_j \mathbf{1}(i) \mathbf{1}(j) \right) \\
    = & \frac{n^2}{m^2} \left( \sum_{i=1}^n x_i^2 y_i^2 \mathbf{1}(i) 
 + \sum_{i\neq j; 
 i,j=1}^n x_i y_i x_j y_j \mathbf{1}(i) \mathbf{1}(j) \right)
\end{align}

\begin{align}
    \mathbf{E}(\langle \hat{x}, \hat{y} \rangle)^2 &  = \left( \frac{n^2}{m^2} \frac{m}{n} \sum_{i=1}^n  x_i^2 y_i^2 \right) + \left( \frac{n^2}{m^2} \frac{m(m-1}{n(n-1)} \sum_{i\neq j; 
 i,j=1}^n x_i y_i x_j y_j \right) \\
 & = \left( \frac{n}{m} \sum_{i=1}^n  x_i^2 y_i^2 \right) + \left( \frac{n}{m} \frac{(m-1)}{(n-1)} \sum_{i\neq j; 
 i,j=1}^n x_i y_i x_j y_j \right)
\end{align}

Thus, the variance, 

\begin{equation}
    \mathbf{V}(\langle \hat{x}, \hat{y} \rangle) =  \left( \left( \frac{n}{m} - 1 \right) \sum_{i=1}^n  x_i^2 y_i^2 \right) + \left( \left( \frac{n}{m} \frac{(m-1)}{(n-1)}  - 1\right)  \sum_{i\neq j; 
 i,j=1}^n x_i y_i x_j y_j \right)
\end{equation}

\begin{equation}
    \mathbf{V}(\langle \hat{x}, \hat{y} \rangle) =  \left( \left( \frac{n}{m} - 1 \right) \sum_{i=1}^n  x_i^2 y_i^2 \right) + \left( \left( \frac{(m-n)}{m(n-1)}\right)  \sum_{i\neq j; 
 i,j=1}^n x_i y_i x_j y_j \right)
\end{equation}

\subsection{Parameter sharing}

the estimator of inner products. Let $h$ be the final position mapping resulting from $P$ and $R$. The signs from $R$ are represented by the function $g$
\begin{equation}
    \langle \hat{x}, \hat{y} \rangle = \sum_{i=1}^m \left( \sum_{j=1}^n x_j g(j) \mathbf{1}(h(j) = i) \right) \left( \sum_{j=1}^n y_j g(j) \mathbf{1}(h(j) = i) \right) 
\end{equation}
where $\mathbf{1}$ is indicator.
The equation can be simplified as,

\begin{equation}
    \langle \hat{x}, \hat{y} \rangle = \sum_{i=1}^m \left( \sum_{j,k=1}^n x_j y_k g(j) g(k) \mathbf{1}(h(j) = i) \mathbf{1}(h(k) = i) \right) 
\end{equation}
which can be further simplified as,

\begin{equation}
    \langle \hat{x}, \hat{y} \rangle = \sum_{i,j=1}^n  x_i y_j g(i) g(j) \mathbf{1}(h(i) = h(j)) 
\end{equation}

\begin{equation}
    \mathbf{E}(\langle \hat{x}, \hat{y} \rangle) = \langle x, y \rangle
\end{equation}

\begin{equation}
    \langle \hat{x}, \hat{y} \rangle^2 = \sum_{a,b,c,d} x_a y_b x_c y_d g(a) g(b) g(c) g(d) \mathbf{1}(h(a) = h(b)) \mathbf{1}(h(c) = h(d))
\end{equation}
Only 4 cases give the non-zero terms in expectations
\begin{itemize}
    \item $a=b=c=d$
    \item $a=b \neq c=d$
    \item $a=c\neq b=d$
    \item $a=d \neq b=c$
\end{itemize}

\begin{align}
    \mathbf{E}(\langle \hat{x}, \hat{y} \rangle^2) & = \sum_{i} x_i^2 y_i^2 \\
    & + \sum_{i\neq j} x_i y_i x_j y_j \\
    & + \sum_{i \neq j} x_i^2 y_j^2 \mathbf{E}(\mathbf{1}(h(i) = h(j))) \\
    & + \sum_{i \neq j} x_i y_i x_j y_j \mathbf{E}(\mathbf{1}(h(i) = h(j)))
\end{align}
Thus variance,
\begin{align}
    \mathbf{V}(\langle \hat{x}, \hat{y} \rangle) & = \sum_{i \neq j} x_i^2 y_j^2 \mathbf{E}(\mathbf{1}(h(i) = h(j))) \\
    & + \sum_{i \neq j} x_i y_i x_j y_j \mathbf{E}(\mathbf{1}(h(i) = h(j)))
\end{align}

\begin{equation}
    \mathbf{Pr}(h(i) = h(j) = \textrm{ Pr(i and j in different chunk) Pr( i and j from different chunks collide ) }
\end{equation}
Assming that $m$ divides $n$
\begin{equation}
    \mathbf{Pr}(h(i) = h(j) = \left( 1 - \frac{m-1}{n- 1} \right) \frac{1}{m} 
\end{equation}

\begin{align}
    \mathbf{V}(\langle \hat{x}, \hat{y} \rangle) & = \frac{1}{m} \frac{n-m}{n-1} \left( \sum_{i \neq j} x_i^2 y_j^2  + \sum_{i \neq j} x_i y_i x_j y_j \right)
\end{align}

\subsection{Norms}
For norms (precisely square of norm) we can just use $x=y$ in the previous analysis

\begin{equation}
    \mathbf{V}_{prune}(\langle \hat{x}, \hat{x} \rangle) =  \left( \left( \frac{n}{m} - 1 \right) \sum_{i=1}^n  x_i^4 \right) + \left( \left( \frac{(m-n)}{m(n-1)}\right)  \sum_{i\neq j; 
 i,j=1}^n x_i^2 x_j^2 \right)
\end{equation}

\begin{align}
    \mathbf{V}_{\roastpp}(\langle \hat{x}, \hat{x} \rangle) & = \frac{2}{m} \frac{n-m}{n-1} \left( \sum_{i \neq j} x_i^2 x_j^2 \right)
\end{align}

\subsection{Comparison}
As is clear from the two expressions, which method of compression is better depends on the $x$ and $y$ at hand. So in order to get a sense of which method is better on ``average", let us take some interesting distributions from which these points are drawn.

\subsection{Average case over different distributions (Inner products) }
Let us assume that each of these $x$ and $y$ are independently drawn from distribution defined below
\begin{equation}
    x_i \sim \mathcal{N}(0, \sigma_i^2)
\end{equation}
In real scenarios each input may have different strengths or importance ( as we will see in the next section ) thus, we assume them to be distinct and of course, we can always get the scenario of equal strengths by setting all equal to a single $\sigma$.
Let us look inner products first,

\begin{equation}
    \mathbf{E}_D( \mathbf{V}_{prune}(\langle \hat{x}, \hat{y} \rangle))  =  \left( \left( \frac{n-m}{m} \right) \sum_i \sigma_i^4 \right)
\end{equation}

\begin{align}
     \mathbf{E}_D(\mathbf{V}_{\roastpp}(\langle \hat{x}, \hat{y} \rangle)) & = \frac{1}{m} \frac{n-m}{n-1} \left( \sum_{i \neq j} \sigma_i^2 \sigma_j^2 \right)
\end{align}

We know that,
\begin{equation}
    \sum_{i \neq j} a_i a_j = \sum_{i < j} 2 a_i a_j  \leq \sum_{i < j} (a_i^2 + a_j^2) = (n-1) \sum a_i^2
\end{equation}

Thus, 
\begin{align}
     \mathbf{E}_D(\mathbf{V}_{\roastpp}(\langle \hat{x}, \hat{y} \rangle)) & \leq \frac{1}{m} \frac{n-m}{n-1} \left( (n-1) \sum_{i} \sigma_i^4 \right) = \mathbf{E}_D( \mathbf{V}_{prune}(\langle \hat{x}, \hat{y} \rangle))
\end{align}

Thus parameter sharing is strictly better than pruning and equality happens only when data is uniformly distributed. ( all standard deviations are same). Whenever there is a slight imbalance in signal strengths {\roastpp} is superior to pruning.

\subsection{Average case over different distributions (Norms) }
Under same distribution,

\begin{align}
    \mathbf{E}_\mathcal{D}(\mathbf{V}_{\roastpp}(\langle \hat{x}, \hat{x} \rangle)) & = \frac{2}{m} \frac{n-m}{n-1} \left( \sum_{i \neq j} \sigma_i^2 \sigma_j^2 \right)
\end{align}

\begin{align}
    \mathbf{E}_\mathcal{D}(\mathbf{V}_{prune}(\langle \hat{x}, \hat{x} \rangle)) & =  \left( \left( \frac{n}{m} - 1 \right) \sum_{i=1}^n  3 \sigma_i^4 \right) + \left( \left( \frac{(m-n)}{m(n-1)}\right)  \sum_{i\neq j; 
 i,j=1}^n \sigma_i^2 \sigma_j^2 \right) \\
 & \geq  \left( \left( \frac{n}{m} - 1 \right)  3 \frac{1}{(n-1)} \sum_{i\neq j} \sigma_i^2 \sigma_j^2\right) + \left( \left( \frac{(m-n)}{m(n-1)}\right)  \sum_{i\neq j; 
 i,j=1}^n \sigma_i^2 \sigma_j^2 \right) \\
 & = \left( \left( 3\frac{n-m}{m(n-1)} \right) \right) + \left( \left( \frac{(m-n)}{m(n-1)}\right) \right)  \left( \sum_{i\neq j; 
 i,j=1}^n \sigma_i^2 \sigma_j^2 \right)  \\
 & = \left( \left( 2\frac{n-m}{m(n-1)} \right) \right) \left( \sum_{i\neq j; 
 i,j=1}^n \sigma_i^2 \sigma_j^2 \right)  \\
 & = \mathbf{E}_\mathcal{D}(\mathbf{V}_{\roastpp}(\langle \hat{x}, \hat{x} \rangle))
\end{align}

Thus,
\begin{equation}
 \mathbf{E}_\mathcal{D}(\mathbf{V}_{prune}(\langle \hat{x}, \hat{x} \rangle)) \geq \mathbf{E}_\mathcal{D}(\mathbf{V}_{\roastpp}(\langle \hat{x}, \hat{x} \rangle))    
\end{equation}

Thus parameter sharing is strictly better than pruning and equality happens only when data is uniformly distributed. ( all standard deviations are same). Whenever there is a slight imbalance in signal strengths {\roastpp} is superior to pruning on average.

\section{Theory : Analysis of linear models under Pruning and {\roastpp}}
Consider the following features $x_1, x_2, ... x_n$ such that,
\begin{equation}
    Var(x_i) = V(x_i) = \sigma_x^2 \; \forall i  \qquad \textrm{ and } \qquad Cov(x_i, x_j) = C(x_i, x_j) = 0 \; \forall i \neq j
\end{equation}

\paragraph{Pruning.}
\begin{align}
    Res(\hat{y}) = Var( y - \sum_i \alpha_i \mathbf{1}(i) x_i) 
\end{align}
\begin{align}
    Res(\hat{y}) = V(y) + \sum_i \alpha_i^2 \mathbf{1}(i) V(x_i) - 2 \sum_i \alpha_i \mathbf{1}(i) C(y,x_i)
\end{align}

Minimizing the residual, we get,
\begin{equation}
    \mathbf{1}(i) \alpha_i = \frac{C(y, x_i)}{\sigma_x^2} 
\end{equation}

\begin{align}
    Res(\hat{y}) = V(y) - \sum_i \frac{C(y,x_i)^2}{\sigma_x^2} \mathbf{1}(i)
\end{align}

\begin{align}
    Res(\hat{y}) = \sigma_y^2 \left(1  - \sum_i \frac{C(y,x_i)^2}{\sigma_y^2 \sigma_x^2} \mathbf{1}(i) \right)
\end{align}

Let $\rho(i) = \frac{Cov(y, x_i)}{\sigma_y \sigma_x}$

\begin{align}
    Res(\hat{y}) = \sigma_y^2 \left(1  - \sum_i \rho(i)^2 \mathbf{1}(i) \right)
\end{align}
\begin{align}
    Res(\hat{y}) = \sigma_y^2 \left(1  - \frac{1}{k} k \sum_i \rho(i)^2 \mathbf{1}(i) \right)
\end{align}
\begin{align}
    Res(\hat{y}) = \sigma_y^2 \left(1  - \frac{1}{k} \langle \hat{\rho}, \hat{\rho} \rangle_{prune} \right)
\end{align}

where $\rho$ is the vector of all correlation coefficients
where $\hat{\rho}$ is the estimation from data compression of $\rho$

\paragraph{\roastpp}
\begin{align}
    Res(\hat{y}) = Var\left( y - \sum_{j=1}^m \alpha_j \sum_i \mathbf{1}(i,j)g_i x_i\right) 
\end{align}
\begin{align}
    Res(\hat{y}) = V(y) + \sum_j \left(\alpha_j^2 V\left(\sum_i \mathbf{1}(i,j)g_i x_i\right) \right) - 2 \sum_j \alpha_j C(y, \sum_i \mathbf{1}(i,j) g_ix_i)
\end{align}

Using the same analysis,
\begin{equation}
    \alpha_j = \frac{C(y, \sum_i \mathbf{1}(i,j) g_ix_i )}{V(\sum_i \mathbf{1}(i,j) g_ix_i)} = \frac{\sum_i \mathbf{1}(i,j) g_i C(y,x_i) }{k \sigma_x^2}
\end{equation}

\begin{align}
    Res(\hat{y}) = V(y) + \sum_j \left(\alpha_j^2 k \sigma_x^2 \right) - 2 \sum_j \alpha_j  k \sigma_x^2 \alpha_j
\end{align}

\begin{align}
    Res(\hat{y}) = V(y) - \sum_j \left(\alpha_j^2 k \sigma_x^2 \right)
\end{align}

\begin{align}
    Res(\hat{y}) = \sigma_y^2 \left( 1  - \sum_j \left(\frac{ \left(\sum_i \mathbf{1}(i,j) g_i C(y,x_i) \right)^2 }{k \sigma_x^2 \sigma_y^2}  \right) \right)
\end{align}

\begin{align}
    Res(\hat{y}) = \sigma_y^2 \left( 1  - \frac{1}{k} \sum_j \left(\left(\sum_i \mathbf{1}(i,j) g_i \frac{C(y,x_i)}{\sigma_x \sigma_y} \right)^2 \right) \right)
\end{align}

\begin{align}
    Res(\hat{y}) = \sigma_y^2 \left( 1  - \frac{1}{k} \sum_j \left(\sum_i \mathbf{1}(i,j) g_i \rho(i) \right)^2  \right)
\end{align}

\begin{align}
    Res(\hat{y}) = \sigma_y^2 \left( 1  - \frac{1}{k} \langle \hat{\rho}, \hat{\rho} \rangle_{\roastpp}  \right)
\end{align}

So the residual is minimized based on how large is the second term.

\subsection{Comparison}
It is clear from the above two analysis that essential question is about norm-preservation of the vector of $(\rho(1), \rho(2), \rho(3) , .... , \rho(n)$ under sampling and {\roastpp} sketching.
The residual in case of no compression can be obtained by just setting $\mathbf{1}(i) = 1$ for all $i$ and hence $k=1$

\begin{equation}
    Res(\hat{y}) = \sigma_y^2 \left( 1  - \langle \rho, \rho \rangle \right) 
\end{equation}

\begin{equation}
    \frac{\textrm{Res}_{comp}(\hat{y}) - \textrm{Res}(\hat{y})}{\textrm{Res}(\hat{y})} =  \frac{\langle \rho , \rho \rangle - \frac{1}{k} \langle \hat{\rho}, \hat{\rho} \rangle }{1 - \langle \rho , \rho \rangle}
\end{equation}

\subsection{Summary}
We now analyse the effectiveness of {\rand} pruning and {\roastpp} in theoretical setting. We first analyse the advantage in dimensionality reduction problem (preserving norms and inner products under data compression). In context of data compression, pruning implies dropping vector components and {\roastpp} implies sketching of vector components under the specific mapping function. We later show that quality of learning of linear models can be reduced to norm preservation  under data compression.
One important thing to note here is that we get these theoretical results only for {\roastpp} mapping and these results would not have been possible if we were analysing {\roast}.

\paragraph{Data Compression}
Consider two vectors $x,y \in R^n$. Let the parameter budget be $m = n/k $. For the sake of simplicity we assume that $k$ is an integer. Let $\hat{x}, \hat{y}$ be the compressed $x$ and $y$. The estimation of inner products is,
\begin{equation}
    \langle \hat{x}, \hat{y} \rangle_{\textrm{prune}} = k  \left( \sum_{j=1}^n x_j y_j \mathbf{1}(j) \right) 
\end{equation}
\begin{equation}
    \langle \hat{x}, \hat{y} \rangle_{\textrm{\roastppsmall}} = \sum_{i=1}^m \left( \sum_{j=1}^n x_j g(j) \mathbf{1}(h(j) = i) \right) \left( \sum_{j=1}^n y_j g(j) \mathbf{1}(h(j) = i) \right)
\end{equation}
where $\mathbf{1}(i)$ is an indicator if $i^{th}$ component is sampled and $\mathbf{1}(h(j) == i)$ is an indicator if the hash mapping $h$ maps $j$ to $i$ and $g(j)$ is the sign hash function. The following theorem shows the theoretical advantage of {\roastpp}. In order to make equations simpler, we assume that the parameters are first permuted before applying the {\roastpp} mapping.

\begin{theorem}
    Given vectors $x,y \in R^n$ and under pruning and \roastpp with memory budget $m = n / k$ for $k \in \mathbf{N}$, the estimates for both methods are unbiased. The variances are as follows,
    
    \begin{equation}
    \mathbf{V}_{\textrm{sample}}(\langle \hat{x}, \hat{y} \rangle_{prune}) =  \left( \left( \frac{n}{m} - 1 \right) \sum_{i=1}^n  x_i^2 y_i^2 \right) + \left( \left( \frac{(m-n)}{m(n-1)}\right)  \sum_{i\neq j; 
 i,j=1}^n x_i y_i x_j y_j \right)
    \end{equation}

    \begin{align}
    \mathbf{V}_{h}(\langle \hat{x}, \hat{y} \rangle_{\roastppsmall}) & = \frac{1}{m} \frac{n-m}{n-1} \left( \sum_{i \neq j} x_i^2 y_j^2  + \sum_{i \neq j} x_i y_i x_j y_j \right)
\end{align}
As can be seen, depending on exact vectors, either method can be better than each other. As both these methods are data agnostic, let us take a look at an average case over some data distribution. Let $x_i$ and $y_i$ be i.i.d drawn from any distribution with $\mu=0, \sigma=\sigma_i$. Let this data distribution be $\mathcal{D}$. Then we have,

\begin{equation}
 \mathbf{E}_\mathcal{D}(\mathbf{V}_{sample}(\langle \hat{x}, \hat{y} \rangle_{
 \textrm{prune}
 })) \geq \mathbf{E}_\mathcal{D}(\mathbf{V}_{h}(\langle \hat{x}, \hat{y} \rangle)_{\roastppsmall}) 
\end{equation}
with equality being achieved if and only if all $\sigma_i$s are equal.
\end{theorem}
We can similarly analyse preservation of norms and the results are summarized below,
\begin{theorem}
    Given vectors $x\in R^n$ and under pruning and \roastpp with memory budget $m = n / k$ for $k \in \mathbf{N}$, the estimates for both methods are unbiased. The variances are as follows,
    
    \begin{equation}
    \mathbf{V}_{\textrm{sample}}(\langle \hat{x}, \hat{x} \rangle_{prune}) =   \left( \left( \frac{n}{m} - 1 \right) \sum_{i=1}^n  x_i^4 \right) + \left( \left( \frac{(m-n)}{m(n-1)}\right)  \sum_{i\neq j; 
 i,j=1}^n x_i^2 x_j^2 \right)
    \end{equation}

    \begin{align}
    \mathbf{V}_{h}(\langle \hat{x}, \hat{x} \rangle_{\roastppsmall}) = \frac{2}{m} \frac{n-m}{n-1} \left( \sum_{i \neq j} x_i^2 x_j^2 \right)
\end{align}
Let $x_i$ and $y_i$ be i.i.d drawn from a distribution with $\mathcal{N}(\mu=0, \sigma=\sigma_i)$. Let this data distribution be $\mathcal{D}$. Then we have,

\begin{equation}
 \mathbf{E}_\mathcal{D}(\mathbf{V}_{sample}(\langle \hat{x}, \hat{y} \rangle_{
 \textrm{prune}
 })) \geq \mathbf{E}_\mathcal{D}(\mathbf{V}_{h}(\langle \hat{x}, \hat{y} \rangle)_{\roastppsmall}) 
\end{equation}
with equality being achieved if and only if all $\sigma_i$s are equal.
\end{theorem}
\textbf{Interpretation:} As pointed out in theorems, the variance really depends on what $x$ (or $x$, $y$) are we considering. So, we want to understand for what portion of points is pruning or {\roastpp} superior. Hence, we take the expectation over data. In case of perfectly balanced distribution (which implies a radial symmetric distribution), the expectations are equal. So exactly same proportions of data points favor pruning / {\roastpp}. However, in practice distribution is hardly symmetric. In fact distributions are generally power-law, In this situation {\roastpp} is strictly better in terms of portions of data-space that it approximates better. The actual advantage will depend on scale of power-law with higher power-law favoring {\roastpp} 

\paragraph{Model Compression}

Now let us take a look at pruning and \roastpp in context of model parameters for linear models. Consider the learning of a linear model for $y$ given signal vector $x_1, ..., x_n$.  Let the correlation of output variable $y$ with each $x_i$ be $\rho_i$. Let the vector $\rho = (\rho_1, \rho_2, ... \rho_n)$. We can then analyse the best residual obtained (i.e. learned model residual) once the pruning sample of weights or mapping of {\roastpp} has been decided. The following result reduces the residual to data compression on the vector $\rho$.

\begin{theorem}
    The optimal (least) residual for linear regression problem for target $y$ with standard deviation $\sigma_y$ and input signals $(x_1, x_2, ..., x_n)$ each of which has standard deviation $\sigma_x$, with correlations $\rho_i = \frac{Cov(y, x_1)}{\sigma_x \sigma_y}$, can be written as,
\begin{equation}
    \frac{\textrm{Res}_{comp}(\hat{y}) - \textrm{Res}(\hat{y})}{\textrm{Res}(\hat{y})} =  \frac{\langle \rho , \rho \rangle - \frac{1}{k} \langle \hat{\rho}, \hat{\rho} \rangle }{1 - \langle \rho , \rho \rangle}
\end{equation}
    where $\rho$ is the vector of all $\rho_i$s and $\langle \hat{\rho}, \hat{\rho} \rangle $ is the estimate of norm of $\rho$ vector under data compression (pruning or {\roastpp})
\end{theorem}

\textbf{Interpretation:} As can be seen from the above theorem, the residual obtained by the compressed model (both pruning and {\roastpp}) depends on how well the norm of the vector $\rho$ is preserved under compression. As we have seen in the previous theorem, if the correlations are not balanced (which is usually the case in real data), then {\roastpp} is better than pruning on average.
\section{Convergence of {\roast} and {\roastpp}} \label{app:convergence}

We consider a simple case of Lipschitz continuous function with gradient descent and look at the stability region for the learning rate.

The update is 
\begin{equation}
    \theta_{t+1} = \theta_t - \eta \nabla F (\theta_t)
\end{equation}
Using the lipschitz smoothness, we have
\begin{equation}
    F(y) \leq F(x) + \langle \nabla F(x), y - x \rangle + \frac{L}{2} || y - x ||^2  
\end{equation}

\begin{equation}
    F(\theta_{t+1}) \leq F(\theta_t) + \langle \nabla F(\theta_t), - \eta \nabla F(\theta_t) \rangle + \eta^2 \frac{L}{2} || \nabla F(\theta_t) ||^2  
\end{equation}

\begin{equation}
    F(\theta_{t+1}) \leq F(\theta_t) - \eta ||\nabla F(\theta_t) ||^2 + \eta^2 \frac{L}{2} || \nabla F(\theta_t) ||^2  
\end{equation}

\begin{equation}
    F(\theta_{t+1}) \leq F(\theta_t) - \eta ||\nabla F(\theta_t) ||^2 ( 1 - \eta \frac{L}{2} )  
\end{equation}
For stability, $\eta < 2 / L$

\paragraph{\roast}
Under parameter sharing we consider the following function  $\Lambda$ is a diagonal matrix with $\Lambda_{ii} = g(i) \lambda_i$
\begin{equation}
    G(\psi) = F(\theta = \Lambda R \psi)
\end{equation}

\begin{align*}    
    || \nabla G(\psi_1) - \nabla G(\psi_2) ||  & = || (\Lambda R)^\top (\nabla F(\theta_1 = \Lambda R \psi_1) - \nabla F(\theta_2 = \Lambda R \psi_2)) || \\
    & \leq \rho(\Lambda R) || (\nabla F(\theta_1 = \Lambda R \psi_1) - \nabla F(\theta_2 = \Lambda R \psi_2)) || \\
    & \leq \rho(\Lambda R) L_f || ( \Lambda R \psi_1 - \Lambda R \psi_2)) || \\
    & \leq \rho^2(\Lambda R) L_f || \psi_1 - \psi_2 ||
\end{align*}
Thus, $L_g = \rho^2(\Lambda R) L_f$

\begin{equation}
\rho^2(\Lambda R) = \max(\textrm{eigen-value}) ( R^\top \Lambda^2 R ) = \max_j \sum_{i=1}^n \lambda_i^2 \mathbf{1}(h(i) = j)     
\end{equation}

Thus the stability range is $\left(0, \frac{2}{L\max_j \sum_{i=1}^n \lambda_i^2 \mathbf{1}(h(i) = j) } \right )$
\paragraph{\roastpp}

Now let us analyse the gradient scaling algorithm. Again, Consider an update of the gradient descent algorithm, we go from $\psi_t$ to $\psi_{t+1}$ in compressed space which corresponds to $\theta_{t} = \Lambda R \psi_{t}$. Where $\Lambda$ and R are as defined above. Also, the gradient scaler is written as a diagonal $m\times m$ matrix $\Gamma$ (this is a different notation than used in main text where it was a vector in $R^m$.)

We have the update equation.
\begin{equation}
    \psi_{t+1}  = \psi_{t}   - \eta \Gamma \nabla  G(\psi_{t})
\end{equation}
Thus, 
\begin{equation}
    \theta_{t+1}  = \theta_{t}   - \eta \left(\Lambda R \Gamma R^\top \Lambda^\top \right) \nabla  F(\theta_{t})
\end{equation}

As $F$ is a L-smooth function, we have,
\begin{equation}
    F(\theta_{t+1}) \leq F(\theta_t) + \langle \nabla F(\theta_t), \theta_{t+1} - \theta_{t} \rangle + \frac{L}{2} || \theta_{t+1} - \theta_{t} ||^2  
\end{equation}

\begin{equation}
    F(\theta_{t+1}) \leq F(\theta_t) - \eta \nabla F(\theta_t)^\top  \left(\Lambda R \Gamma R^\top \Lambda^\top \right) \nabla  F(\theta_{t}) + \frac{\eta^2 L}{2}  \nabla F(\theta_t)^\top \left(\Lambda R \Gamma R^\top \Lambda^\top \right)^\top \left(\Lambda R \Gamma R^\top \Lambda^\top \right) \nabla F(\theta_t) 
\end{equation}

\begin{equation}
    F(\theta_{t+1}) \leq F(\theta_t) - \eta \nabla F(\theta_t)^\top \left(  \left(\Lambda R \Gamma R^\top \Lambda^\top \right)  - \frac{\eta L}{2}  \left(\Lambda R \Gamma R^\top \Lambda^\top \right)^\top \left(\Lambda R \Gamma R^\top \Lambda^\top \right) \right) \nabla F(\theta_t) 
\end{equation}

Consider the matrix,
\begin{equation}
M = \left(  \left(\Lambda R \Gamma R^\top \Lambda^\top \right)  - \frac{\eta L}{2}  \left(\Lambda R \Gamma R^\top \Lambda^\top \right)^\top \left(\Lambda R \Gamma R^\top \Lambda^\top \right) \right)    
\end{equation}
The matrix $\left(\Lambda R \Gamma R^\top \Lambda^\top \right)$
 is a $n\times n$ diagonal matrix with $\left(\Lambda R \Gamma R^\top \Lambda^\top \right)_{ii} = \lambda^2_i \gamma_{h(i)}$
Thus matrix $\left(\Lambda R \Gamma R^\top \Lambda^\top \right)^\top \left(\Lambda R \Gamma R^\top \Lambda^\top \right)$ is also diagonal matrix with $\left(\left(\Lambda R \Gamma R^\top \Lambda^\top \right)^\top \left(\Lambda R \Gamma R^\top \Lambda^\top \right)\right)_{ii} = \lambda^4_i \gamma_{h(i)}^2$. 
Thus diagonal element of the matrix in above equation is 
\begin{equation}
    M_{ii} = \lambda_i^2 \gamma_{h(i)}\left(1  - \frac{\eta L}{2} \lambda_i^2 \gamma_{h(i)} \right)
\end{equation}

If $\gamma_{j} = \frac{1}{\sum_{i,h(i) = j} \lambda_i^2}$, and if $\eta < 2 / L$,
then, 
\begin{equation}
    \forall i, M_{ii} > 0
\end{equation}
as all $\lambda_i > 0$
Since M is a diagonal matrix, M is a positive  definite and thus $F(\theta_{t+1}) < F(\theta_t)$ and is stable. Note that the $\gamma$ chosen here is pessimistic (assumes that other $\lambda$'s can be arbitrarily close to 0 and thus performs aggressive scaling. In practice we find that effective update scaler performs well.

\section{Pareto Continuity} \label{app:mapping}

The load factor of each bucket is essentially the number of partitions the original flattened out vector is divided into since each partition adds exactly one element to each bucket. Thus the load $l$ for each bucket is 
\begin{equation}
  l(i) \leq \left\lceil \frac{n}{m} \right\rceil  
\end{equation}
Which is optimal.  When $n=m$, the number of collisions is $0$ and thus RPS model and full model are equivalent (except for memory layout of the weights). 

The cache fetches can increase if a roast chunk is separated while partitioning. The maximum number of separations is - first it is separated while partitioning and then each of the chunk goes through wrap around (which breaks it into two) - thus $4$. This means maximum additional of $3$ cache line fetches that are potentially partially filled. 
\section{Experimental Details}\label{sec:app_experimentdetails}

\textbf{Learning Rate Schedules} We choose the learning rate schedule given by \citet{tanaka2020pruning} for the milestone steps. We find the the learning rate given in \citet{tanaka2020pruning} is not the best learning rate even if we use there code base as a starting point of our implementation. So we run the baseline full models with varying learning rate from $\{0.001, 0.01, 0.1, 1, 10\}$ with lr drop-rate of $0.1$ and schedules that are mentioned in table \ref{tab:hyperparameters}.  We find that $0.1$ works better for all the full model trainings. 

We use the same learning rate schedules and other common hyperparmeters for {\roastpp} and pruning methods.

\begin{table}[h]
\centering
\caption{hyperparameters for both models RESNET20 and VGG11 }\label{tab:hyperparameters}
\begin{tabular}{|c|c|c|c|}
\hline
\textbf{Hyperparameter}     & \textbf{CIFAR-10} & \textbf{CIFAR-100} & \textbf{TINY-IMAGENET} \\ \hline
\textbf{Base Learning Rate} & 0.1               & 0.1                & 0.1                    \\ \hline
\textbf{Milestones}         & 80,120            & 60,120             & 30,60,80               \\ \hline
\textbf{Learning rate drop} & 0.1               & 0.1                & 0.1                    \\ \hline
\textbf{batch size}         & 128               & 128                & 128                    \\ \hline
\textbf{total epochs}       & 160               & 160                & 100                    \\ \hline
\end{tabular}
\end{table}

\paragraph{{\roastpp} hyperparameters.} The standard deviation for the RPS array is set to 0.01 for RESNET20 model and 0.05 for VGG11 model. The choice is made following the rule of thumb that scale factors should be in check. However, with gradient scaling, it is not required to be very strict about these parameters

\paragraph{Pruning iterations.} We use 100 iterations for snip and synflow which has been shown to improve its performance. For mag and rand, the number of iterations don't really matter. For GrasP, we tried 100 pruning iterations but that turns out to be bad for the pruning. We thus choose to do only 1 iteration (single shot) for grasp as is originally done. 

\paragraph{Compression Details.} For all the methods and models, we \textbf{do not} compress the bias and batchnorm parameters (henceforth called as bias\_parameters) into compression. The compression rate is applied to rest of the parameters. The compression rate thus can be computed as,
\begin{equation}
    compression = \frac{total - bias\_parameters}{compressed - bias\_parameters}
\end{equation}

\paragraph{LTR.} we use open lth repository for this experiment where each at each level, we remove 0.25 fraction of the parameters using global masking.
\section{Theoretical Scaler vs. Effective udpate scaler} \label{app:scaling_comparison}
In the paper, we show two scaling mechanisms. We find that our initial implemnetation of \textbf{theoretical-scaler} had a bug which caused it to perform worse than \textbf{effective-update-scaler}. For consistency all the results in main paper are presented using \textbf{effective-update-scaler}. We present the comparison among two scalers here. We find that, in fact, \textbf{theoretical-scaler} performs better.

\begin{figure}[h]
    \centering
    \begin{subfigure}{0.48\textwidth}
    \includegraphics[scale=0.4]{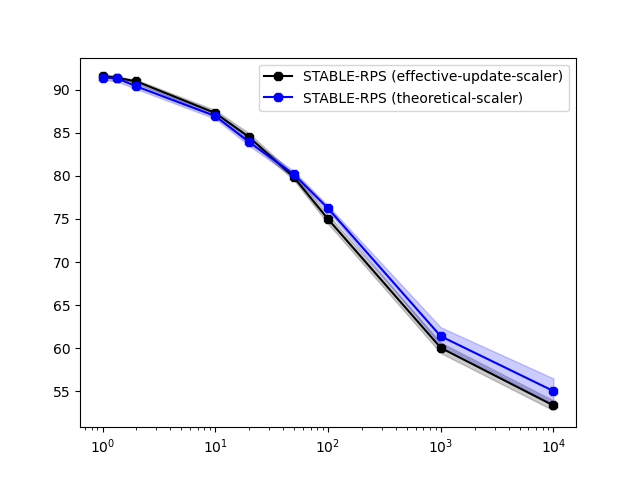}
    \caption{CIFAR-10(RESNET20)}
    \end{subfigure}
    \begin{subfigure}{0.48\textwidth}\includegraphics[scale=0.4]{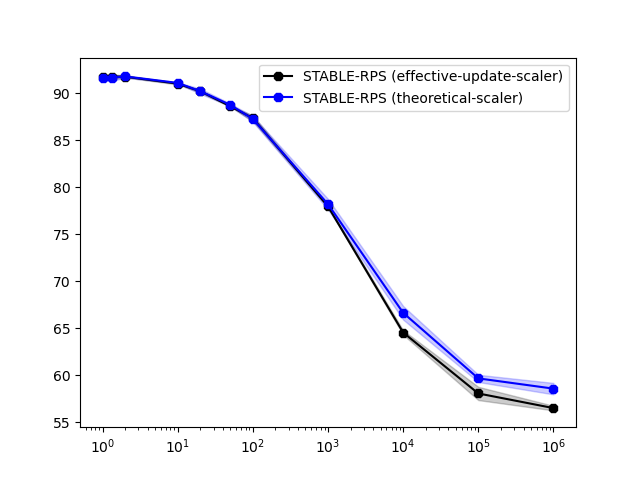}
    \caption{CIFAR-10(VGG11)}
    \end{subfigure}
    \centering
    \begin{subfigure}{0.48\textwidth}
    \includegraphics[scale=0.4]{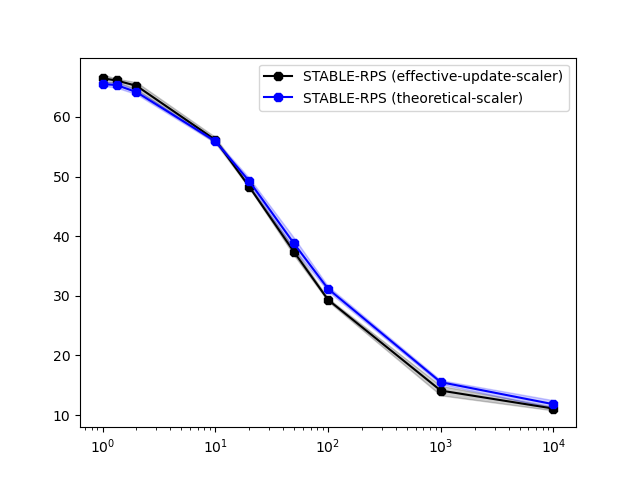}
    \caption{CIFAR-100(RESNET20)}
    \end{subfigure}
    \begin{subfigure}{0.48\textwidth}\includegraphics[scale=0.4]{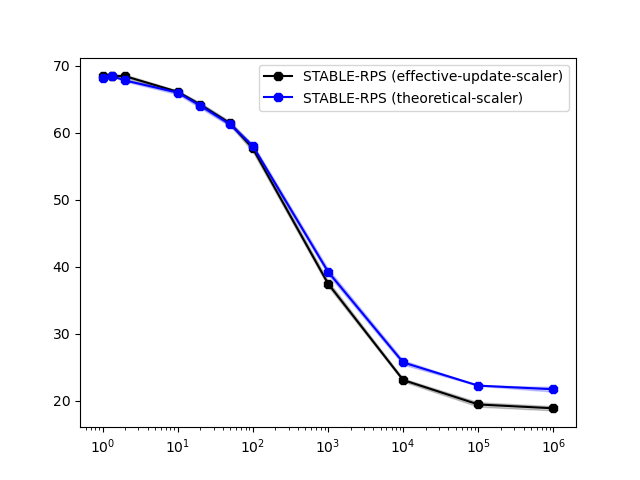}
    \caption{CIFAR-100(VGG11)}
    \label{fig:enter-label}    
    \end{subfigure}
    \centering
    \begin{subfigure}{0.48\textwidth}
    \includegraphics[scale=0.4]{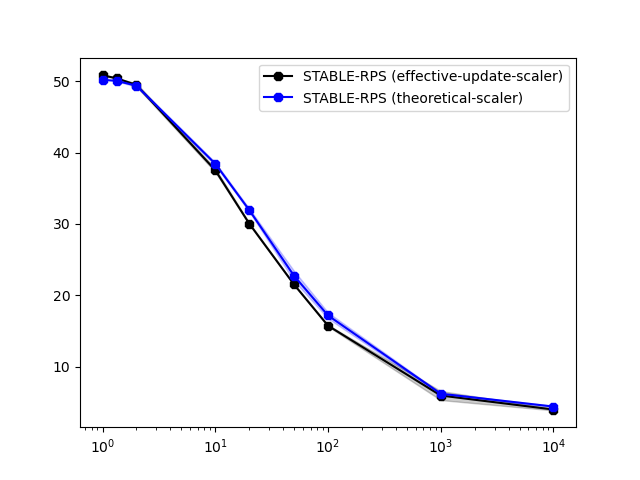}
    \caption{TINYIMAGE-NET(RESNET20)}
    \end{subfigure}
    \begin{subfigure}{0.48\textwidth}\includegraphics[scale=0.4]{scaler_comparison/cifar10-vgg11.png}
    \caption{TINYIMAGE-NET(VGG11)}
    \end{subfigure}

\end{figure}
%\section{Analysis : Norms of solutions obtained from different methods}

\end{document}